%% file: main.tex
\documentclass{article}

\usepackage[whole]{bxcjkjatype}
\usepackage{bm}
\usepackage{amstext}
\usepackage{ulem}
\usepackage{algorithm}
\usepackage{algorithmic}
\usepackage{color}
\usepackage{amsmath}
\usepackage{multirow}
\usepackage{multicol}
\usepackage{diagbox}
\usepackage{tabularx}
\usepackage{arxiv}

\usepackage[utf8]{inputenc} % allow utf-8 input
\usepackage[T1]{fontenc}    % use 8-bit T1 fonts
\usepackage{hyperref}       % hyperlinks
\usepackage{url}            % simple URL typesetting
\usepackage{booktabs}       % professional-quality tables
\usepackage{amsfonts}       % blackboard math symbols
\usepackage{nicefrac}       % compact symbols for 1/2, etc.
\usepackage{microtype}      % microtypography
\usepackage{cleveref}       % smart cross-referencing
\usepackage{lipsum}         % Can be removed after putting your text content
\usepackage{graphicx}
\usepackage{natbib}
\usepackage{doi}

\title{HiPerformer: Hierarchically Permutation-Equivariant Transformer for Time Series Forecasting}

\author{
    Ryo Umagami \\
    The University of Tokyo \\
    \texttt{umagami@mi.t.u-tokyo.ac.jp}
    \And
    Yu Ono \\
    The University of Tokyo \\
    \texttt{ono@mi.t.u-tokyo.ac.jp}
    \And
    Yusuke Mukuta \\
    The University of Tokyo / RIKEN AIP \\
    \texttt{mukuta@mi.t.u-tokyo.ac.jp}
    \And
    Tatsuya Harada \\
    The University of Tokyo / RIKEN AIP \\
    \texttt{harada@mi.t.u-tokyo.ac.jp}
}

\renewcommand{\shorttitle}{HiPerformer: Hierarchically Permutation-Equivariant Transformer}

\hypersetup{
pdftitle={HiPerformer},
pdfauthor={Ryo Umagami, Yu Ono, Yusuke Mukuta, Tatsuya Harada},
pdfkeywords={Permutation-Equivariance, Time Series Forecasting, Multi-Agent Trajectory Prediction, Hierarchical Forecasting},
}

\begin{document}
\maketitle

\input{section/0_abstract}
\input{section/1_introduction}
\input{section/2_relatedworks.tex}
\input{section/3_methods}
\input{section/4_experiments.tex}
\input{section/5_conclusion.tex}
\input{section/6_acknowledgements.tex}
\bibliographystyle{unsrtnat}
\bibliography{references}

\end{document}

%% file: section/0_abstract.tex
\begin{abstract}

% 実世界に存在する時系列には、互いに強く影響しあっている系列同士があると知られている~\citep{sukcharoen2016dependence}。特に、企業が業種で分類される様に、系列がクラス分けされている場合には、クラス同士も影響しあっていることがあり、系列間の関係は階層的な依存構造をなしているといえる。そこで本研究では、時系列の集合を入力として、それに含まれる系列間の階層的な依存構造を捕捉可能な時系列予測モデルを提案する。提案モデルは、本研究で定義した階層的な順序共変性という性質をもっており、任意のサイズの集合に適用可能である。これにより、多くの既存手法が抱えている、系列の新規参入や退場が生じるデータに適用できないという問題を解決している。実世界のデータでの実験を通して、提案手法がstate of the artの手法を上回っていることを示した。

% 複数時系列予測において系列間の影響を上手く取り出すことが重要である
% 特に株価のように、しばしば構成要素は同じ性質を持ったグループに分かれており、このグループ構造に矛盾しない影響を取り出すモデルにより有効に予測が行えると期待される
% そこでこのグループ構造を考慮したモデルを設計するうえで、グループ内、グループ毎の構成要素のインデクス入れ替えに着目した階層的な順序共変性の概念を提案する
% 予測モデルが階層的な順序共変性を持てば予測は構成要素のグループ関係と矛盾なく行える
% 我々はこの階層的な順序共変性を持つモデルとして、同じグループの構成要素同士の関係と、グループ同士の関係の両方を考慮する予測モデルを提案する
% 実世界のデータでの実験により、提案手法がstate of the artの手法を上回っていることを示した
It is imperative to discern the relationships between multiple time series for accurate forecasting.
In particular, for stock prices, components are often divided into groups with the same characteristics, and a model that extracts relationships consistent with this group structure should be effective.
Thus, we propose the concept of hierarchical permutation-equivariance, focusing on index swapping of components within and among groups, to design a model that considers this group structure.
When the prediction model has hierarchical permutation-equivariance, the prediction is consistent with the group relationships of the components.
Therefore, we propose a hierarchically permutation-equivariant model that considers both the relationship among components in the same group and the relationship among groups.
The experiments conducted on real-world data demonstrate that the proposed method outperforms existing state-of-the-art methods.

\end{abstract}

\keywords{Permutation-Equivariance \and Time Series Forecasting \and Multi-Agent Trajectory Prediction \and Hierarchical Forecasting}

%% file: section/1_introduction.tex
\section{Introduction} \label{sec:introduction}

% 時系列データ予測は、生活のさまざまな場面で人々の意思決定を助ける。たとえば商品の売上個数が予測できれば、過不足のない発注により利益を最大化できる。実世界には無数の時系列が存在するが、その中には互いに強く影響しあっている系列同士がある。たとえば米国株式においては、同じセクターに所属する銘柄の間には依存関係および強い相関が見られることを示されている~\citep{sukcharoen2016dependence}。
Time series forecasting facilitates decision-making in several facets of life. If the number of units of a product sold can be predicted, profits can be maximized by placing orders without excess or shortage. In practice, there are numerous time series, some of which are strongly influenced by each other. Previous studies have demonstrated strong correlations and dependencies among stocks belonging to the same sector in the U.S. stock market~\citep{sukcharoen2016dependence}.

% 系列同士が影響しあっているのなら、その集合をモデルに同時に入力し、その関係性を考慮した予測をするのが合理的である。その場合、系列の新規参入や退場を扱えることが要求される。これは小売なら新商品の発売と廃盤に、株価ならIPOと倒産に対応する。このとき、系列の入力順序に基づいて特定の系列同士の依存関係を獲得する枠組みでは、この課題は解決されない。なぜならば、たとえば新規参入した系列に新たなindexが割り振られたとしたときに、その新しいindexの系列と、もともと存在していた系列との関係を獲得する機構は学習されていないからだ。しかし実際には、系列間の関係を活用しようとするほとんどの既存手法では、系列の個数と入力順序を固定して、その順序を使って学習しており、系列の入退場に対応することができない。この問題を解決するのは順序共変性~\citep{deepsets}という性質である。順序共変な機構とは、系列の集合を入力として受けつけ、入力系列の順序を変えると出力の順序もそのまま変わるような機構のことである。詳しくは\ref{sec:permutation}で述べる。つまり、時系列の集合について予測をする場合、系列間の関係を獲得可能であり、サイズ可変の系列集合を扱うことのできる順序共変なモデルが望ましい。
If time series are influencing each other, it is reasonable to input the set of series into the model simultaneously and make forecasts considering their relationships. 
Then the model must be equipped to accommodate the entry and exit of the series. The entry and exit of a series correspond to the launch and discontinuation of new products in retail and initial public offerings and bankruptcies in stock prices. 
However, a model that acquires dependencies among specific series based on the input order of the series cannot solve this problem.
This is because the model lacks a mechanism for acquiring the relationship between the new index and the existing series.
But in fact, most existing methods that leverage the relationships among series cannot handle the entry and exit of series, because they fix the number of series and the input order, and use that order to train a model.
Permutation-equivariant models are used to address this issue ~\citep{deepsets}. 
A permutation-equivariant mechanism takes a set of series as input and outputs a corresponding series, that is rearranged according to the permutation of the input. 
Details are given in \ref{sec:permutation}.
In summary, when predicting a set of time series, we must have a permutation-equivariant model that can acquire relationships among the series and handle variable-size series sets.

% 実世界の時系列群はしばしば、スポーツにおけるチーム、企業における業種のように、同じ性質を持った要素をまとめたグループに分かれている。
% 本研究では各グループのことをクラスと呼ぶことにする。
% 時系列予測においてこのクラス構造を利用して系列感の影響を学習することで有効に予測できると期待される。
% 本研究ではこのようなクラス関係を考慮した予測モデルの構築指針として、クラス関係を考慮した系列要素間の入れ替えに着目し、階層的な順序共変性という概念を提案する。
% また、このような階層的な順序共変性を満たす予測モデルとして時間方向、クラス内、クラス間の影響をself-attentionでモデル化したhierarchical permutation-equivariant transformer (HiPerformer)を提案する
% 種々の実時系列を用いた予測実験によりhierarchical permutation equivariant transformerは既存モデルより良い予測性能を示した
Real-world time series groups are often divided into classes that group together elements with the same characteristics, such as teams in sports or industries in businesses.
It is expected that this class structure could be effective in time series forecasting by learning the dependencies among series.
We propose the concept of hierarchical permutation-equivariance as a guideline for building prediction models considering such class relationships.
This concept focuses on the permutation of series considering the class relationships.
Furthermore, we propose a hierarchically permutation-equivariant transformer (HiPerformer) that models intra-class, inter-class and time effects with self-attention and is hierarchically permutation-equivariant.
HiPerformer outperforms existing models in experiments using various real time series.

Our contributions are as follows:
\begin{itemize}
    \item We define hierarchical permutation and hierarchical permutation-equivariance.
    \item We proposed two types of models that can accommodate the entry and exit of a series.
        \begin{enumerate}
            \item HiPerformer:\:A model which is hierarchically permutation-equivariant and can capture the hierarchical dependencies among series. This model requires class information for each series.
            \item HiPerformer-w/o-class:\:A model which is permutation-equivariant and capture the dependencies among series. This model does not require class information for each series.
        \end{enumerate}
    \item We proposed a distribution estimator that outputs joint distributions with time-varying covariance to handle uncertain real time series.
    \item The proposed model outperforms state-of-the-art methods in experiments on artificial and real datasets where the series have a hierarchical dependency structure. This shows that utilizing the hierarchical dependency structure is effective for forecasting time series that are classified.
\end{itemize}

%% file: section/2_relatedworks.tex
\section{Related Works}

\subsection{Time Series Forecasting} \label{sec:ts}
% Autoregressive Model (AR)などの従来の線形時系列予測モデルは、表現力に乏しい上に、トレンドや季節性などのドメイン知識を手動で入力する必要があった。これらの問題を解決するために、さまざまなdeep neural network modelsが提案されている。Recurrent Neural Network(RNN)~\citep{DBLP:journals/corr/ChoMGBSB14}は、内部に状態をもつようなニューラルネットワークであり、理論上過去の全入力を考慮できるはずであるが、勾配消失や勾配爆発の問題から長期間の情報を出力に反映することは難しい。Long Short-Term Memory (LSTM)や、Gated Recurrent Unit　(GRU)はこれらの問題を緩和させるが、依然として遠い過去の情報を獲得することはできないとされている~\citep{khandelwal2018sharp}。また、推論時には過去の推論結果を予測に使用するため、長期間の予測では推論誤差が蓄積してしまうという問題がある。近年提案されたTransformer~\citep{vaswani2017}は、attention-mechanismによって、過去の情報に距離に関係なくアクセスすることを可能にし、より長期間の依存関係を獲得できるようになった。計算量削減や季節性の考慮などの点においてTransformerを改良したものがたくさん発表されている~\citep{li2019enhancing, zhou2021informer, wu2021autoformer, zhou2022fedformer, liu2021pyraformer}。
Conventional linear time series prediction models, such as the autoregressive model (AR) are slightly expressive and require manual input of domain knowledge such as trends and seasonality. 
To address these issues, various deep neural network models have been proposed, such as recurrent neural network (RNN)~\citep{DBLP:journals/corr/ChoMGBSB14}.
RNNs, though theoretically can consider all prior inputs, suffer from the challenges of gradient vanishing and explosion, impeding their use for long-term information reflection. 
Long short-term memory (LSTM) and gated recurrent units (GRUs) alleviate these problems but still cannot retrieve information in the distant past~\citep{khandelwal2018sharp}.
Furthermore, since past inference results are used for prediction during inference, inference errors can accumulate in long-term predictions. 
The recently proposed Transformer~\citep{vaswani2017} allows access to past information regardless of distance through the attention mechanism, allowing for longer-term dependencies to be obtained. 
Numerous improvements to Transformer have been published for computational savings and seasonality considerations~\citep{li2019enhancing, zhou2021informer, wu2021autoformer, zhou2022fedformer, liu2021pyraformer}.

\subsection{Capturing Relationships Among Series} \label{sec:CapturingRelationships}
% \ref{sec:introduction}で説明したような系列間の関係を考慮可能なモデルとしては、伝統的なものではVector Autoregressive Model (VAR)があり、これをRNNによって深層学習に拡張したものがDeepVAR~\citep{salinas2019high}である。DeepVARはmultivariate time seriesが従う分布のパラメータを求めることができる確率的な予測手法である。TransformerとGenerative Adversarial Networks (GAN)~\citep{goodfellow2020generative}を組み合わせたAST~\citep{wu2020adversarial}では、同時に達成されるものとして尤もらしい各系列の予測を出力することができる。系列をノードに、系列間の関係をエッジに対応させて、Graph Neural Network~\citep{scarselli2008graph}を使う手法もある。AGCRN~\citep{bai2020adaptive}は、RNNとGraph Convolutional Networks~\citep{defferrard2016convolutional, kipf2016semi}を組み合わせた手法である。\\
% 系列間の関係を考慮することが有効なタスクとして、互いに影響し合う複数のエージェントの動きを予測する、Multi-Agent Trajectory Predictionがある。Multi-Agent Trajectory Predictionは、自動運転~\citep{zhao2021tnt, salzmann2020trajectron++, 2012.01526, leon2021review}やrobot planning~\citep{kretzschmar2014learning, schmerling2018multimodal}、スポーツの解析~\citep{felsen2017will}などの多くの応用先があり、盛んに研究されている。GRIN~\citep{li2021grin}は、Graph Attention Networks~\citep{velickovic2017graph}を用いることでエージェント間の高度なinteractionを獲得するのに成功している。\\
% 系列間の階層的な関係性に着目したタスクとして、Hierarchical Forecastingがある。このタスクでは上位階層の系列が下位階層の系列の集約であり、bottom階層の系列だけでなく、上位の集約階層の系列も含めて予測をする。たとえばこの設定に該当するデータの例としては、ある学校の生徒数は全ての学年の生徒数の和であり、ある学年の生徒数は全てのクラスの生徒数の和である、というようなものがある。HierE2E~\citep{rangapuram2021end}やDPMN~\citep{olivares2021probabilistic}は、和に関する階層的な一貫性を予測に持たせることで、各系列を個別に予測するよりも精度が高くなる場合があると示している。\\
% 以上の手法は各分野において良い結果を残しているものの、系列間の階層的な依存関係を考慮した手法は存在せず、情報を損失しているといえる。
The vector autoregressive model (VAR) is the traditional model for examining series relationships, and DeepVAR is an extension of this model to incorporate deep learning with RNNs. 
DeepVAR is a probabilistic forecasting method that can determine the parameters of the distribution that target multivariate time series follow.
AST~\citep{wu2020adversarial} combining Transformers and generative adversarial networks (GANs)~\citep{goodfellow2020generative} can output predictions for each series that are viable for simultaneous achievement. 
Another approach is using GNNs~\citep{scarselli2008graph}, mapping series to nodes and relationships between series to edges. 
AGCRN~\citep{bai2020adaptive} is a method that combines RNNs and graph convolutional networks~\citep{defferrard2016convolutional, kipf2016semi}.\\
Multi-agent trajectory prediction is predicting the movements of multiple agents that influence each other, and it is effective to consider the relationships among the agents.
Multi-agent trajectory prediction has various applications, such as automated driving~\citep{zhao2021tnt, salzmann2020trajectron++, 2012.01526, leon2021review}, robot planning~\citep{kretzschmar2014learning, schmerling2018multimodal}, sports analysis~\citep{felsen2017will}, and is being actively researched.
GRIN~\citep{li2021grin} has been successful in obtaining advanced interactions between agents using GATs~\citep{velickovic2017graph}.\\
Hierarchical forecasting focuses on hierarchical relationships among series.
In this task, a series in the upper hierarchy is an aggregate of series in the lower hierarchy, and forecasts are made for the series in the bottom hierarchy and the series in the upper aggregation hierarchy. 
In this setting, data on the number of students in a school is the aggregate of the student totals in all grades, and the student count in each grade is the sum of the student numbers in all classes.
HierE2E~\citep{rangapuram2021end} and DPMN~\citep{olivares2021probabilistic} indicate that hierarchical consistency regarding sums may lead to higher accuracy than forecasting each series separately.\\
Despite their successes, none of the previous methods consider hierarchical dependencies among series, leading to an information loss.

\subsection{Permutation-Equivariance / Invariance} \label{sec:permutation}
% 順序共変性と順序不変性は~\citep{deepsets}で定義された。順序共変な機構$f$とは、行列$X = [\bm{x}_1, ..., \bm{x}_N]^T$と任意の置換行列$P_{\pi} \in \mathbb{R}^{N \times N}$に対し$f(P_{\pi}X) = P_{\pi}f(X)$が成り立つもののことである。これに対して順序不変な機構$f$とは、$f(P_{\pi}X) = f(X)$が成り立つもののことである。\\
% ~\citep{lee2019}では、系列集合を入力として、他の系列情報を考慮した特徴量を出力する機構として、Set Attention Block(SAB), Induced Set Attention Block(ISAB), Pooling by Multi-Head Attention(PMA)が提案された。SABは、系列集合に対して、系列間でSelf Attentionを取る手法である。系列の数を$N$とすると、SABの計算量は$\mathcal{O}(N^2)$となり、系列数が大きい場合に計算量が爆発してしまう。これを解決するのがISABであり、SABとほぼ同等の性能を持ちながら、計算量がハイパーパラメータ$m$を使って$\mathcal{O}(mN)$となり、$N$に対して線形に抑えることができる。PMAは、系列集合を系列方向に圧縮する、順序不変な機構である。圧縮では、平均値をとるAverage Poolingや、最大値をとるMax Poolingなどが使われることが多いが、PMAはAttention機構を用いているためより柔軟なPoolingが可能である。計算量はISABと同じくハイパーパラメータと系列数の積に対して線形である。注目すべきは、SAB, ISAB, PMAは任意のサイズの系列集合を入力として受けつけ、系列方向に順序共変または順序不変であり、系列間の関係を考慮した特徴抽出が可能だということである。これらの性質は本研究の目的と相性が良いため、提案モデルの基礎となっている。
Permutation-equivariance and permutation-invariance were defined in ~\citep{deepsets}. A permutation-equivariant mechanism $f$ is; for a tensor $X = [\bm{x}_1, . \bm{x}_N]^T$ and any substitution matrix $P_{\pi} \in \mathbb{R}^{N \times N}$, $f(P_{\pi}X) = P_{\pi}f(X)$ holds. On the other hand, a permutation-invariant mechanism $f$ is one for which $f(P_{\pi}X) = f(X)$ holds. \\

In ~\citep{lee2019}, set attention block (SAB) and induced set attention block (ISAB) were proposed which is permutation-equivariant and take self-attention among a set of series.
Self-attention can acquire relationships among series and is used in Transformer introduced also in \ref{sec:ts}.
The computational complexity of SAB is on a square order to the number of series, while ISAB exhibits comparable performance with a linear complexity.
Also, pooling by multi-head attention (PMA) was proposed which is permutation-invariant and compress a series set in the series dimension using multi-head attention.
Notably, SAB, ISAB, and PMA accept an arbitrary size set of series as input, are permutation-equivariant or permutation-invariant for series dimension, and can extract features considering the relationships among series. 
These properties align with the objectives of this study and form the basis of the proposed model.

%% file: section/3_methods.tex
\section{Methods} \label{sec:methods}

\subsection{Problem Definition}
% 本研究におけるタスクは、$S$系列の、時刻$T-T_{in}:T$の間の$D_{in}$個の説明変数$X \in \mathbb{R}^{S \times T_{in} \times D_{in}}$と各系列のクラス情報$\bm{c}\in \mathbb{N}^S$から、それに続く時刻$T+1:T+T_{out}$の間の$D_{out}$個の目的変数$Y \in \mathbb{R}^{S \times T_{out} \times D_{out}}$が従う、時変な確率分布を求めることである。ここでクラス情報$\bm{c}$について、$\bm{c}_i=\bm{c}_j$となるときに$i$番目の系列と$j$番目の系列が同じクラスに所属することを表す。\\
% まず入力テンソル$X$を、クラス情報$\bm{c}$を用いてクラス分けする。$S$個の系列が$C$クラスに分かれており、最も所属系列の多いクラスの系列数を$S_c$とすると、$X$は、$\mathbb{R}^{C \times S_c \times T_{in} \times D_{in}}$のテンソルに整形することができる。これを改めて$X$と置く。このとき$X_{i,j}$は$i$番目のクラスの$j$番目の系列の情報を表す。ただし、$j$が$i$番目のクラスに所属する系列数を超えている$X_{i,j}$については、存在しない系列の情報なので参照されないようにする。これ以降$\bm{c}$が使われることはなく、特徴量としてモデルへ入力されることもない。\\
Suppose that we have a collection of $S$ related multivariate time series $\{X_{i, \: T-T_{in}+1:T}\}_{i=1}^S$, where $X_{i, \: T-T_{in}+1:T} \in \mathbb{R}^{T_{in} \times D_{in}}$ denotes $D_{in}$ explanatory variables of time series $i$ during time $T-T_{in}+1:T$. Furthermore, we have class information of each series $\bm{c}\in \mathbb{N}^S$, when $\bm{c}_i=\bm{c}_j$, it means time series $i$ and time series $j$ belong to the same class.
We will predict the time-varying probability distribution that $\{Y_{i, \: T+1:T+T_{out}}\}_{i=1}^S$ follows, where $Y_{i, \: T+1:T+T_{out}} \in \mathbb{R}^{T_{out} \times D_{out}}$ denotes $D_{out}$ objective variables of time series $i$ during time $T+1:T+T_{out}$.
Series influence each other hierarchically, and the acquisition of hierarchical dependencies is effective in forecasting.

\subsection{Hierarchical Permutation-Equivariance} \label{sec:hierperm}
\begin{figure}[t]
\vskip 0.2in
\begin{center}
\includegraphics[width=0.7\columnwidth]{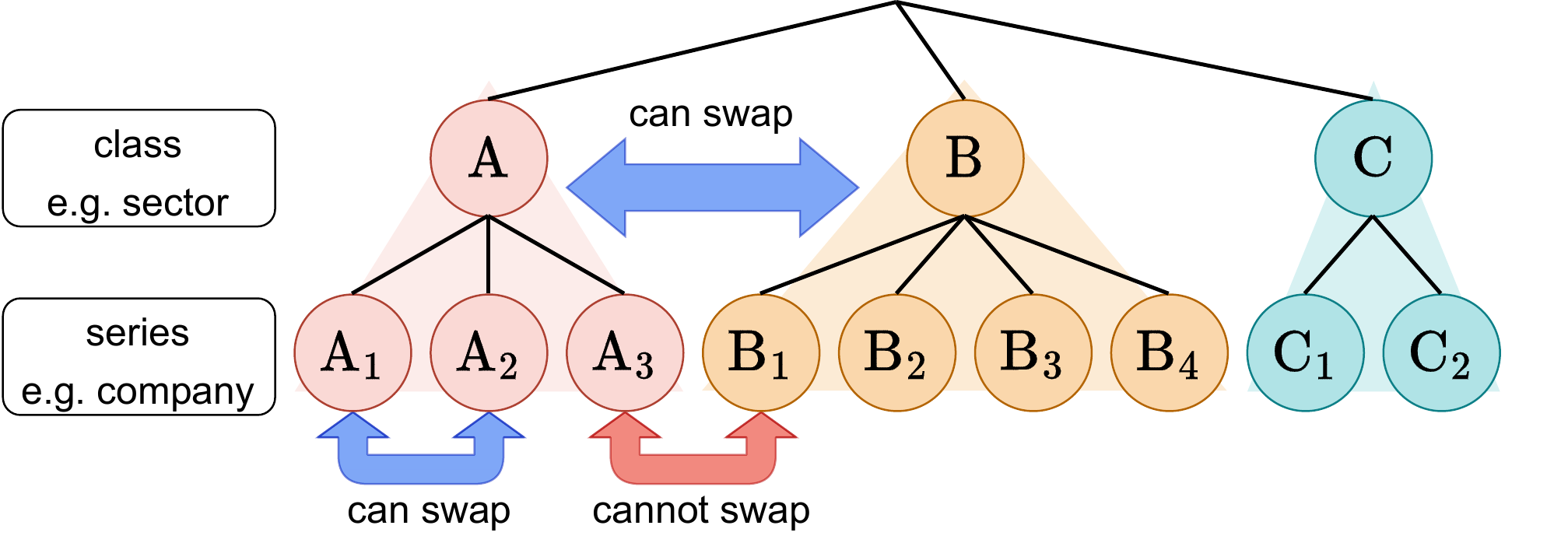}
\caption{Hierarchical permutation}
\label{fig:hier}
\end{center}
\vskip -0.2in
\end{figure}

% Hierarchically-Permutation-Equivarianceは、\ref{sec:permutation}のPermutation-Equivarianceを拡張した枠組みである。まず前段階として新たに、階層的な順序変換を導入する。階層的な順序変換とは、次の(1)または(2)を満たす変換のことと定義する: (1)クラスが同じ系列の入れ替え、(2)クラス自体の入れ替え。つまり階層的な順序変換とは、クラスが違う系列同士の入れ替えを禁止するような、系列の入れ替えということだ。たとえば、クラスA, Bにわかれた系列集合$\{\text{A}_1, \text{A}_2, \text{B}_1, \text{B}_2, \text{B}_3\}$がこの順で与えられた場合を考える。このとき(1)の変換は、$\{\text{A}_2, \text{A}_1, \text{B}_3, \text{B}_1, \text{B}_2\}$のような変換であり、(2)の変換は$\{ \text{B}_1, \text{B}_2, \text{B}_3, \text{A}_1, \text{A}_2\}$のような変換である。(1)と(2)を合わせた$\{\text{B}_3, \text{B}_1, \text{B}_2, \text{A}_2, \text{A}_1\}$もまた階層的な順序変換である。そして、階層的な順序変換に対して順序共変なことを、階層的に順序共変である、と定義する。
% 階層的に順序共変なモデルは、クラスの入力順序、クラスが同じ系列の入力順序によらない予測が可能であり、もとのクラス構造に矛盾しない予測を行うことができる。ゆえに本研究では、階層的に順序共変であり、系列間の階層的な依存構造を捕捉可能な予測モデルを提案する。\todo{階層的な順序変換の図}
% 階層的に順序共変なモデルは、クラスの入力順序、クラスが同じ系列の入力順序によらない予測が可能であり、系列間の階層的な依存関係を獲得するうえで情報損失が少ない。ゆえに本研究では、階層的に順序共変であり、系列間の階層的な依存構造を捕捉可能な予測モデルを提案する。\todo{階層的な順序変換の図}
``Hierarchical permutation-equivariance'' is a paradigm extension of permutation-equivariance described in \ref{sec:permutation}. 
Preliminarily, we introduce ``hierarchical permutation''. 
A hierarchical permutation is one that satisfies (1) or (2): (1) a permutation of the series in the same class, and (2) a permutation of the class itself. 
Specifically, hierarchical permutations are permutations except for permutations among series in different classes.
For example, consider the case where series sets $\{\text{A}_1, \text{A}_2, \text{B}_1, \text{B}_2, \text{B}_3\}$, divided into classes A and B, are given in this order. 
Here, the permutation in (1) is one like $\{ \text{A}_2, \text{A}_1, \text{B}_3, \text{B}_1, \text{B}_2\}$ and the permutation in (2) is like $\{ \text{B}_1, \text{B}_2, \text{B}_3, \text {A}_1, \text{A}_2\}$.
Combining (1) and (2), $\{\text{B}_3, \text{B}_1, \text{B}_2, \text{A}_2, \text{A}_1\}$ is also a hierarchical permutation. 
\ref{fig:hier} visualizes what is a hierarchical-permutation and what is not.
We then define ``hierarchically permutation-equivariant'' as being equivariant for hierarchical permutation of series.
Predictions of a hierarchically permutation-equivariant model are: independent of the input order of classes and series in the same class, and consistent with the class structure.
Therefore, we propose a prediction model that is hierarchically permutation-equivariant and can capture the hierarchical dependencies among series.

%%%%%%%%%%%%%%%%%%%%%%%%%
%%%%%%%%%%%%%%%%%%%%%%%%%

%%%%%%%%%%%%%%%%%%%%%%%%%
\begin{figure}[t]
\vskip 0.2in
\begin{center}
\includegraphics[width=0.7\columnwidth]{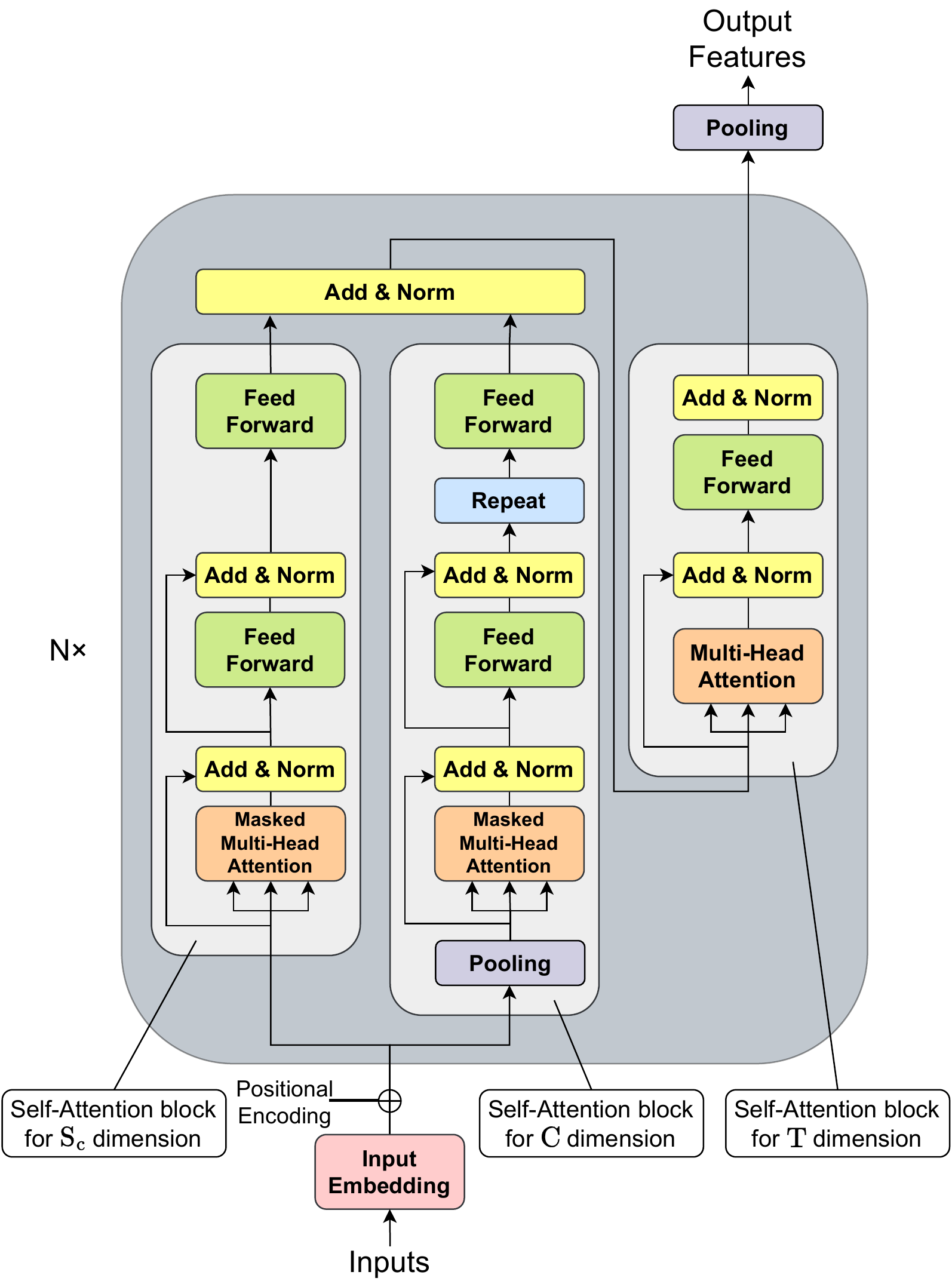}
\caption{Feature extractor architecture}
\label{fig:model_architecture}
\end{center}
\vskip -0.2in
\end{figure}
%%%%%%%%%%%%%%%%%%%%%%%%%

%%%%%%%%%%%%%%%%%%%%%%%%%
\begin{algorithm}[t]
   \caption{3D self-attention}
   \label{alg:3dselfatt}
\begin{algorithmic}[1]
   \STATE {\bfseries Input:} $M \in \mathbb{R}^{C \times S_c \times T_{in} \times D_{model}}$

   \STATE
   
   \STATE {\textcolor{blue}{/*
   \;Self-Attention Block for $S_c$ Dimension\\ 
   \;\;\;\: Curriculate $M^{S_c} \in \mathbb{R}^{C \times S_c \times T_{in} \times D_{model}}$\;*/}}
   \FOR{$c=1$ {\bfseries to} $C$}
        \FOR{$t=1$ {\bfseries to} $T_{in}$}
            \STATE $M^{S_c}_{c, \: :, \: t, \: :} \leftarrow \text{SelfAttForSc}(M_{c, \: :, \: t, \: :})$
        \ENDFOR
   \ENDFOR
   
   \STATE

   \STATE \textcolor{blue}{/*
   \;Self-Attention Block for $C$ Dimension\\ 
   \;\;\;\: Curriculate $M^C \in \mathbb{R}^{C \times S_c \times T_{in} \times D_{model}}$\;*/}
   \FOR{$c=1$ {\bfseries to} $C$}
        \STATE $M^P_{c, \: :, \: :, \: :} \leftarrow \text{Pool}(M_{c, \: :, \: :, \: :}) \in \mathbb{R}^{T_{in} \times D_{model}}$
   \ENDFOR
   % \STATE $M^P \in \mathbb{R}^{C \times T_{in} \times D_{model}}$
   \FOR{$t=1$ {\bfseries to} $T_{in}$}
       \STATE $M^P_{:, \: t, \: :} \leftarrow \text{SelfAttForC}(M^P_{:, \: t, \: :})$
   \ENDFOR
   
   \FOR{$s=1$ {\bfseries to} $S_c$}
        \STATE $M^C_{:, \: s, \: :, \: :} \leftarrow M^P$
   \ENDFOR
   
   \STATE

   \STATE \textcolor{blue}{/*
   \;Self-Attention Block for $T$ Dimension\\ 
   \;\;\;\: Curriculate $M^T \in \mathbb{R}^{C \times S_c \times T_{in} \times D_{model}}$\;*/}
   \STATE $M^T \leftarrow M^{S_c} + M^C$
   \FOR{$c=1$ {\bfseries to} $C$}
        \FOR{$s=1$ {\bfseries to} $S_c$}
            \STATE $M_{c, \: s, \: :, \: :} \leftarrow \text{SelfAttForT}(M_{c, \: s, \: :, \: :})$
        \ENDFOR
   \ENDFOR

   \STATE
   
   \STATE {\bfseries Output:} $M^T \in \mathbb{R}^{C \times S_c \times T_{in} \times D_{model}}$
\end{algorithmic}
\end{algorithm}
%%%%%%%%%%%%%%%%%%%%%%%%%

% First, we classify the input tensor $X$ using the class information $\bm{c}$.
% If $S$ series is divided into $C$ classes and the number of series in the largest class is $S_c$, $X$ can be formatted into a tensor of $\mathbb{R}^{C \times S_c \times T_{in} \times D_{in}}$. 
% Here, $X_{i,j}$ represents the observation of the series $j$ in class $i$.
% Note that $X_{i,j}$, where $j$ exceeds the number of series belonging to the class $i$, will not be referenced because it is information on a series that does not exist. 
% After this, $\bm{c}$ is never used and never input to the model as a feature.

% 以上の処理を経た入力テンソル$X$を、本研究が提案するFeature ExtractorとDistribution Estimatorで順に処理することで、$Y$の従う分布を求めることができる。これらの提案モジュールは、以下のような性質を持っている。
% \begin{itemize}
%     \item 系列間の階層的な依存構造を捕捉可能である
%     \item 階層的に順序共変であり、任意のサイズの系列集合を入力として受け付けるため、系列の入退場に対応可能である
%     \item 時変で多様な同時分布を出力する
% \end{itemize}
\subsection{Model Overview}
We can obtain the distribution that $Y$ follows by passing the input tensor $X$ through the Feature Extractor and Distribution Estimator proposed in this study.
These proposed modules have the following properties.
\begin{itemize}
    \item Capture hierarchical dependencies among series.
    \item Hierarchically permutation-equivariant and capable of handling the entry and exit of series accepting series set of arbitrary size.
    \item Outputs time-varying and diverse joint distributions.
\end{itemize}

\subsection{Feature Extractor} \label{sec:featureextractor}
% Feature Extractorは、\ref{fig:model_architecture}に示すような、3D Self Attention Layerを直列に積み上げたものである。整形後の$X$はEmbeddingによりモデル内テンソル$M \in \mathbb{R}^{C \times S_c \times T_{in} \times D_{model}}$に変換され、3D Self Attention Layerに入力される。3D Self Attention Layerはクラス内の系列方向($S_c$方向)、クラス方向($C$方向)、時間方向($T$方向)にSelf Attentionをとるblockから構成される。$S_c$方向と$C$方向のSelf Attentionは並列してとられ、その後に$T$方向のSelf Attention Blockが続く。$S_c$方向、$C$方向のSelf Attentionを組み合わせることで系列間の階層的な依存関係を獲得し、最後に$T$方向へのSelf Attentionをとることによって、入力全体を考慮した複雑な時間発展を捉えることができる。さらに、3D Self Attention Layer全体が階層的に順序共変であり、任意のサイズの系列集合を入力として受け付ける。以下で各ブロックについて詳細に説明する。
The feature extractor is a stack of 3D self-attention layers shown in \ref{fig:model_architecture}.
Before passing to it, we classify the input tensor $X$ using the class information $\bm{c}$.
If $S$ series are divided into $C$ classes and the number of series in the largest class is $S_c$, $X$ can be formatted into a tensor of $\mathbb{R}^{C \times S_c \times T_{in} \times D_{in}}$. 
Here, $X_{i,j}$ represents the observation of the series $j$ in class $i$.
Note that $X_{i,j}$, where $j$ exceeds the number of series belonging to the class $i$, will not be referenced because it is information on a series that does not exist. 
After this, $\bm{c}$ is never used and never input to the model as a feature.\\
Then, $X$ is embedded into the in-model tensor $M \in \mathbb{R}^{C \times S_c \times T_{in} \times D_{model}}$, and then it is input into the 3D self-attention layer.

\subsubsection{3D Self-Attention Layer} \label{sec:3dselfatt}
\textbf{Algorithm.}
A 3D self-attention layer consists of blocks that take self-attention for intra-class ($S_c$), inter-class ($C$), and time ($T$) dimensions.
The blocks for $S_c$ and $C$ are parallel, followed by the $T$ block.
The algorithm is shown in \ref{alg:3dselfatt}, and line 3-8, 10-19, 21-27 corresponds $S_c$, $C$, and $T$ dimension block respectively.
SelfAttForSc, SelfAttForC, and SelfAttForT are the self-attention functions for $S_c$, $C$, and $T$ dimensions respectively.
When taking self-attention for $S_c$ dimension, we fix the class index and time, and then take self-attention among series in the class.
When taking self-attention for $C$ dimension, first we compress class information (line 8-10), and then take self-attention among classes with fixing time (line 12-14). Finally, we repeat the self-attention output for $S_c$ dimension (line 15-17).
When taking self-attention for $T$ dimension, we fix the class and series index, and then take self-attention among series in the class.

\textbf{Properties.}
We can obtain hierarchical dependencies among the series by combining self-attention for $S_c$ and $C$ dimensions, and finally, by taking self-attention for $T$, we can capture complex time evolution considering the entire input. 
Furthermore, the entire 3D self-attention layer is hierarchically permutation-equivariant and accepts any size set of series as input. An explanation for this is provided below.

\uline{Self-Attention Block for $S_c$ Dimension.}
 This block is permutation-equivariant for $C$ dimension and accepts any number of classes as input since the same process is applied to all the classes.
As $S_c$ dimension is only subjected to self-attention, it is permutation-equivariant and can accept any number of series from the same class as input. 

\uline{Self-Attention Block for $C$ Dimension.}
This block is permutation-equivariant for $S_c$ and accepts any number of series in a class as input since it only performs permutation-invariant compression and repeats for the $S_c$ dimension.
Also, for the $C$ dimension, since it only considers self-attention, it is permutation-equivariant and accepts any number of classes as input.

\uline{Self-Attention Block for $T$ Dimension.}
This block is hierarchically permutation-equivariant and accepts any number of classes and series in a class as input because the same process is applied to all classes and series in a class for this calculation.

\subsubsection{Post Process and Summary} \label{sec:postsummary}
% 3D Self Attention Layerでモデル内テンソル$M$の次元は変化せず、$M$の1,2次元目をまとめてからゼロパディングされた部分を削除したのちに、$T$方向にPoolingすることにより、各系列の時変な特徴量$Z = (Z_1, \dots , Z_S) \in \mathbb{R}^{S \times T_{out} \times D_{model}}$を得る。$Z$は、3D Self Attentionにより、系列間の階層的な依存構造と系列内の時間発展の両方を捉えた特徴量となっている。\\
% なお、3D Self Attention Layerから\ref{sec:attforc}のSelf Attention Block for $C$ dimensionを削除し、すべての系列が同じクラスに属するとみなすことで、クラス情報を用いない、もう一つの提案モデルになる。こちらのモデルは、階層的ではない単なる順序共変性をもち、系列間の関係を補足可能であり、その他の性質はフルモデルと同様である。
The shape of tensor $M$ does not change in the 3D self-attention layer.
After fusing $C$ and $S_c$ dimensions of $M$, we obtain the time-varying features for each series $Z = (Z_1, \dots, Z_S)\in \mathbb{R}^{S \times T_{out} \times D_{ model}}$ by removing the padded portions and pooling for $T$ dimension.
$Z$ is a feature that captures both the heterarchical dependencies among series and the temporal evolution of each series through 3D self-attention.
Note that by removing the self-attention block for $C$ dimension from the 3D self-attention layer and considering all series in the same class, it becomes another proposed model HiPerformer-w/o-class that does not use class information. 
This model has a mere permutation-equivariance (not hierarchical) and can acquire relationships among series, and has the same properties as the full model.

\subsection{Distribution Estimator} \label{sec:distributionestimator}
% Distribution Estimatorは、\ref{sec:featureextractor}で得た特徴量$Z$を入力として、目的変数$Y$が従う、時変な確率分布を求める機構である。本研究では特に、$Y_{:, \:t, \:d}$が$\mathcal{N}(\mu_{:, t, d}, \Sigma_{:, \::, \:t, \:d})$に従うと仮定し、その平均$\mu \in \mathbb{R} ^ {S \times T_{out} \times D_{out}}$および共分散行列$\Sigma \in \mathbb{R} ^ {S \times S \times T_{out} \times D_{out}}$を出力する。以下で$\mu, \Sigma$の計算アルゴリズムを説明する。
A distribution estimator is a mechanism that takes a feature $Z$ obtained with \ref{sec:featureextractor} as input and predicts a time-varying probability distribution that the target variable $Y$ follows. 
In particular, this study assumes that $Y_{:, \:t, \:d}$ follows a multidimensional normal distribution $\mathcal{N}(\mu_{:, t, d}, \Sigma_{:, \::, \:t, \:d})$, and output its mean $\mu \in \mathbb{R} ^ {S \times T_{out} \times D_{out}}$ and covariance matrix $\Sigma \in \mathbb{R} ^ {S \times S \times T_{out} \times D_{out}}$.
The algorithm for calculating $\mu,\Sigma$ is described below.

\subsubsection{Calculation of $\mu$}
% $\mu$は、$f_{\mu}:\mathbb{R}^{D_{model}} \to \mathbb{R}^{D_{out}}$の線形層により、以下の式のように計算される。
$\mu$ is calculated with a linear layer of $f_{\mu}:\mathbb{R}^{D_{model}} \to \mathbb{R}^{D_{out}}$, as in the following formula.
\begin{equation}
    \mu = (f_{\mu}(Z_1), ..., f_{\mu}(Z_S))^T \in \mathbb{R} ^ {S \times T_{out} \times D_{out}} \label{calcmu}
\end{equation}

\subsubsection{Calculation of $\Sigma$} \label{sec:cov}
% 共分散行列は正定値行列であるため、ここで出力する$\Sigma_{:, \::, \:t, \:d}$も正定値でなければならない。
% $f_{R}:\mathbb{R}^{D_{model}} \to \mathbb{R}^{D_{R}}, \; f_{L}:\mathbb{R}^{D_{model}} \to \mathbb{R}^{D_{L}}, \; f_{\sigma}:\mathbb{R}^{D_{model}} \to \mathbb{R}^{D_{\sigma}}$の線形層をそれぞれ$D_{out}$個用意し、$d$番目をそれぞれ$f_{R_d}, f_{L_d}, f_{\sigma_d}$とすると、$\Sigma_{i,j, t, d}$は以下の式のように計算される。
Since the covariance matrix is a positive (semi)definite, the output $\Sigma_{:, \::, \:t, \:d}$ here must also be positive definite.
We have $D_{out}$ linear layers each of $f_{r}:\mathbb{R}^{D_{model}} \to \mathbb{R}^{D_{r}}, \; f_{l}:\mathbb{R}^{D_{model}} \to \mathbb{R}^{D_{l}}, \; f_{\sigma}:\mathbb{R}^{D_{model}} \to \mathbb{R}^{D_{\sigma}}$. 
Let the $d$-th ones be $f_{r_d}$, $f_{l_d}$, and $f_{\sigma_d}$ respectively, then $\Sigma_{i,j, t, d}$ is calculated as in the following equation.
\begin{equation}
    \begin{split}
        \Sigma_{i, \:j, \:t, \:d} &= (r_{i, \: j, \: t, \: d} + l_{i, \: j, \: t, \: d}) \sigma_{i, \: j, \: t, \: d} \\
        where \; r_{i, \: j, \: t, \: d} &= k_{RBF}\left(f_{r_d}(Z_{i, \:t, \::}), f_{r_d}(Z_{j, \:t, \::})\right) \\
        l_{i, \: j, \: t, \: d} &= k_{LIN}\left(f_{l_d}(Z_{i, \:t, \::}), f_{l_d}(Z_{j, \:t, \::})\right) \\
        \sigma_{i, \: j, \: t, \: d} &= k_{LIN}\left(f_{\sigma_d}(Z_{i, \:t, \::}), f_{\sigma_d}(Z_{j, \:t, \::})\right)
    \end{split}
\end{equation}
% ただし、$k_{RBF}(\bm{x}, \bm{y}) = \exp (\gamma \| \bm{x} - \bm{y} \| ^ 2), \; k_{LIN}(\bm{x}, \bm{y}) = \bm{x}^T \bm{y}$であり、それぞれRBFカーネルと線形カーネルを表し、$\odot$は要素積を表す。$\Sigma_{:, \::, \:t, \:d}$は正定値カーネルを和と積で組み合わせたカーネルのグラム行列であるため、正定値行列である。RBFカーネルは非線形な関係を抽出可能であるが値域が$(0, 1]$に制限されてしまうため、線形カーネルを組み合わせることによって表現力を向上させている。
Here, $k_{RBF}(\bm{x}, \bm{y}) = \exp (\gamma \| \bm{x} - \bm{y} \| ^ 2), \; k_{LIN}(\bm{x}, \bm{y}) = \bm{x}^T \bm{y}$, representing the RBF and linear kernels respectively.
$\Sigma_{:, \::, \:t, \:d}$ is positive definite because it is a Gram matrix of kernels combining positive definite kernels by sum and product.
The RBF kernel can extract nonlinear relationships, but its value range is limited to $(0, 1]$. Hence, it is combined with a linear kernel to improve its expressive power.

\subsection{Training Strategy}
% モデルの最適化は、尤度$\sum_{t}\sum_{d} \mathcal{N}(Y_{:, t, d} | \mu_{:, t, d}, \Sigma_{:, \::, \:t, \:d})$を最大化するように、すなわち負の対数尤度$\mathcal{L} = \sum_{t}\sum_{d} -\log \mathcal{N}(Y_{:, t, d} | \mu_{:, t, d}, \Sigma_{:, \::, \:t, \:d})$を最小化するように行われる。ここで、$\bm{u}, \bm{v} \in \mathbb{R}^{D}, W \in \mathbb{R}^{D \times D}$に対して$-\log \mathcal{N}(\bm{u} | \bm{v}, W)$は以下のように計算される。
We optimize the model to maximize the likelihood $\sum_{t}\sum_{d} \mathcal{N}(Y_{:, t, d} | \mu_{:, t, d}, \Sigma_{:, \::, \:t, \:d})$; that is, we can minimize the following negative log likelihood.
\begin{equation}
    \mathcal{L} = \sum_{t}\sum_{d} -\log \mathcal{N}(Y_{:, t, d} | \mu_{:, t, d}, \Sigma_{:, \::, \:t, \:d})
\end{equation}
Note that the negative log-likelihood of a multivariate normal distribution, $-\log \mathcal{N}(\bm{u} | \bm{v}, W)$  is calculated as follows.
\begin{align}
    -&\log \mathcal{N}(\bm{u} | \bm{v}, W) \\
    &= \frac{1}{2} \left[ D \log (2 \pi) + \log |W| + (\bm{u} - \bm{v})^T W^{-1} (\bm{u} - \bm{v}) \right]
\end{align}

%% file: section/4_experiments.tex
\section{Experiments} \label{sec:experiments}
In this section, we validate the performance of the proposed method on two different tasks described in \ref{sec:CapturingRelationships}, multi-agent trajectory prediction, and hierarchical forecasting. For each task, first, we introduce datasets, evaluation metrics, and methods. Thereafter, we conduct evaluations.

\subsection{Multi-Agent Trajectory Prediction}\label{sec:ex_multiagent}

\subsubsection{Datasets}
% 系列間が互いに影響しあっている、一つの人工データと一つの実データを用いる。\\
% \textbf{Charged Dataset}~\citep{kipf2018neural}は、単純な物理法則で制御される荷電粒子の動きを記録した人工データである。各シーンには粒子が5個あり、各粒子は正または負の電荷を持つ。同じ属性の電荷をもつ粒子同士は互いに反発しあい、異なる電荷を持つ粒子同士は互いに引きつけ合う。80期間の軌跡と速度から、続く20期間の軌跡を予測するタスクである。train, validation, testそれぞれについて50K, 10K, 10Kシーンのデータを用いる。提案モデルでは、各粒子の電荷の正負でクラス分けしている。\\
% \textbf{NBA Dataset}~\citep{yue2014learning}は、2012-2013年シーズンのNBAの試合のデータセットである。各シーンには10人の選手と1個のボールがある。40期間の軌跡と速度から、続く10期間の軌跡を予測するタスクである。train, validation, testそれぞれについて80000, 48299, 13464シーンのデータを用いる。提案モデルでは、ボールかどうか、どちらのチームに所属するかの3クラスに分けた。
We use one artificial dataset and one real-world dataset, where the series interact with each other. 

\textbf{Charged Dataset}~\citep{kipf2018neural} is an artificial dataset that records the motion of charged particles controlled by simple physical laws. Each scene contains five particles, each with a positive or negative charge. Particles with the same charge repel each other, and particles with different charges attract each other. The task is to predict the trajectory for the next 20 periods based on the trajectory and velocity for 80 periods. Subsequently, 50K, 10K, and 10K scenes of data are used for training, validation, and testing, respectively. In the proposed model, each particle is classified according to whether its charge is positive or negative.

\textbf{NBA Dataset}~\citep{yue2014learning} is a dataset of National Basketball Association (NBA) games for the 2012--2013 season. Each scene has 10 players and 1 ball and the task is to predict the trajectory of the following 10 periods based on the historical trajectory and velocity for 40 periods. Here, 80000, 48299, and 13464 scenes of data are used for training, validation, and testing, respectively. The proposed model categorizes each series into three classes: ball, and the team to which it belongs.

\subsubsection{Evaluation Metrics} \label{sec:multieval}
% 確率的な予測の精度を評価するため、点推定の結果だけでなく予測分布の尤もらしさも評価する。
To evaluate the accuracy of probabilistic forecasts, we evaluate the likelihood of the forecast distribution and the results of point forecasts. 

\textbf{Average displacement error (ADE) and final displacement error (FDE)}: ADE is the root mean squared error between the ground truth and predicted trajectories. FDE measures the root mean squared error between the ground truth final destination and the predicted final destination. For stochastic models, we report the \textbf{minimum} (and mean for one method) displacement error of all predicted trajectories/destinations. The lower values are preferred.

% \textbf{Negative Log Likelihood (NLL)}: ground truthが、予測された分布に対してどれだけ尤もらしいかを測る指標である。For stochastic models, we report the \textbf{mean} NLL of all  trajectories. We only report this metric on models that are trained with NLL or evidence lower bound (ELBO) ~\citep{sohn2015learning}. Lower means better.
\textbf{Negative log likelihood (NLL)}: It measures how plausible the ground truth is for the predicted distribution. For stochastic models, we report the \textbf{mean} NLL of all trajectories. We only report this metric on models that are trained with NLL or evidence lower bound (ELBO) ~\citep{sohn2015learning}. The lower values are preferred.

\subsubsection{Methods}
% エージェント間のインタラクションを活用している以下の4つの手法をbaselineとして採用する。\\
% \textbf{Fuzzy Query Attention (FQA)}~\citep{kamra2020multi}は、系列の相対的な動き、意図、相互作用から誘導バイアスを獲得するモジュールとRNNを組み合わせた手法である。\\
% \textbf{Neural Relational Inference (NRI)}~\citep{kipf2018neural}は、過去の軌跡から、エージェント間のインタラクションを潜在的なグラフとして表し、variational auto-encoder (VAE) ~\citep{kingma2013auto}の形式で未来の軌跡を生成するものである。\\
% \textbf{Social-GAN}~\citep{gupta2018social}は、マルチモダリティを扱うことを得意としているGAN-basedな伝統的な手法である。\\
% \textbf{GRIN}~\citep{li2021grin}は、各エージェントの意図と、エージェント間の関係を分離した潜在表現を抽出し、Graph Attention Networksと組み合わせることでエージェント間の高度なinteractionを考慮した軌跡の生成を実現している。知る限りこのタスクにおけるstate-of-the-artである。\\
% これらの手法について、特に重要なポイントでの提案手法との比較を、\ref{tab:methodscompare}で示している。
We adopt the following four methods　as baselines that utilize interactions among agents.

\textbf{Fuzzy query attention (FQA)}~\citep{kamra2020multi} is a method that combines RNN with a module that acquires induced bias from the relative movements, intentions, and interactions of sequences. 

\textbf{Neural relational inference (NRI)}~\citep{kipf2018neural} represents interactions among agents as a potential graph from past trajectories and generates future trajectories as variational auto-encoders (VAEs)~\citep{kingma2013auto}. 

\textbf{Social-GAN}~\citep{gupta2018social} is a traditional GAN-based method that effectively handles multimodality. 

\textbf{GRIN}~\citep{li2021grin} extracts latent representations that separate the intentions of each agent and the relationships between agents and generates trajectories considering high-level interactions between agents utilizing GATs~\citep{velickovic2017graph}. To the best of our knowledge, this is the state-of-the-art method in this dataset.

% これらのbaselinesと比較するのは、フルの提案モデルであるHiPerformer-prob、HiPerformer-probから\ref{sec:cov}で説明した共分散の推定を除き軌跡のみを予測する様にしたHiPerformer-det、HiPerformer-probから\ref{sec:attforc}の$C$方向へのSelf Attention Blockを除いたHiPerformer-w/o-class-prob、HiPerformer-w/o-class-probから共分散推定を除いたHiPerformer-w/o-class-detの4つである。HiPerformer-probとHiPerformer-w/o-class-probがNLLをlossとして学習されるのに対し、HiPerformer-detとHiPerformer-w/o-class-detはMAEやMSEなどの距離誤差をlossとして学習される。今回のタスクではMAEをlossとした。また、HiPerformer-w/o-class-probとHiPerformer-w/o-class-detは\ref{sec:featureextractor}で述べた通り、クラス情報を用いないことに注意する。

Next we explain about our proposed model HiPerformer.

\textbf{HiPerformer.}
We compare aforesaid baselines with HiPerformer-prob, the full proposed model; HiPerformer-det, which is HiPerformer-prob without the covariance estimation described in \ref{sec:cov} and only predicts trajectories; HiPerformer-w/o-class-prob, which is HiPerformer-prob without the self-attention block for $C$ dimension in \ref{sec:3dselfatt}; and HiPerformer-w/o-class-det, which is HiPerformer-w/o-class-prob without the covariance estimation.
While HiPerformer-prob and HiPerformer-w/o-class-prob are trained with NLL, HiPerformer-det, and HiPerformer-w/o-class-det are trained with distance errors, such as mean absolute error (MAE) and mean square error (MSE). In this task, we use MAE. Also note that we never use class information for HiPerformer-w/o-class-prob and HiPerformer-w/o-class-det, as described in \ref{sec:postsummary}.
We use SAB described in \ref{sec:permutation} when taking self-attention for $S_c, C$ dimension. 

A comparison of baselines with the proposed method is shown in \ref{tab:methodscompare}.

\textbf{Implementation Details.}
We implement our model using PyTorch ~\citep{paszke2019pytorch}.
We stack two 3D self-attention layer, $D_{model}$ is 64, the number of heads for multi-head attention is 4, the batch size is 8 for Charged dataset and 64 for NBA dataset.
We train our model using Adam optimizer ~\citep{kingma2014adam} with learning rate $1 \times 10^{-3}$.
The results of baseline methods are from ~\citep{li2021grin} except for GRIN.
We train GRIN with parameters used in ~\citep{li2021grin}.

\input{table/methods_compare.tex}

\input{table/multiagent_result.tex}

% \subsubsection{Quantitative Evaluation}
\subsubsection{Quantitative Evaluation}
% \textbf{Baseline Comparison.} ベースラインと提案手法の結果を\ref{tab:multiagent}に示している。モデル名の最後に"min"がついている手法は、予測を確率的に生成する手法で有り、ADEとFDEについてはいくつか生成したもののうちbestなmetricを掲載している。NLLについては、平均を報告している。生成数はChargedで100、NBAで20, 100とした。この指標の計算の仕方は生成系の手法の評価としては一般的であるが~\citep{gupta2018social, mohamed2020social, li2021grin}、FQAや提案手法などの非生成系の手法と実用上の性能を比較する場合、生成的な手法の評価値はbestな値ではなく、平均や中央値などが公平であることを強調しておく。そこでbestな結果が最も良かったGRINについて、生成された予測のmetricの平均値を"GRIN mean"として報告している。\\
% この指標の計算の仕方は生成系の手法の評価としては一般的であるが、FQAや提案手法などの非生成系の手法での指標と単純に比較することはできない。
% ChargedのADEとFDEでは、GRIN minが最高の性能であったが、HiPerformer-w/o-class-prob, HiPerformer-w/o-class-det, HiPerformer-detが肉薄する性能を示した。GRIN meanの結果はすべての提案手法で上回った。NLLでは、提案手法がGRINの1/30程度のNLLを記録し、大きく上回る結果となった。この実験では、クラス情報を用いて階層的な依存構造を獲得しようとするHiPerformerよりも、クラス情報を用いず各系列間の関係を獲得するHiPerformer-w/o-classの方が良い性能であった。これは、粒子の数が5個と少ないうえに関係性が実世界の実世界のものよりも単純であり、他のクラスの粒子の情報を\ref{alg:attforc}の2-3行目のように圧縮するよりも、圧縮されていない生のデータをそのまま用いた方がrichな情報を獲得できたからだと思われる。\\
% NBAについては、すべての指標において提案手法がbaselinesを上回るか同等の性能であった。特に性能が良かったのがHiPerformer-detとHiPerformer-probであり、HiPerformer-probはGRINの1/20以下のNLLを記録した。Chargedのときと異なり、HiPerformerがHiPerformer-w/o-classの性能を上回ったのは、NBAの選手は当然他の選手の所属チームを意識してバスケットボールをプレーしており、チームによりクラス分けしたときの階層的な依存構造がよく実態を捉えているからだと考えられる。HiPerformerがその階層構造をうまく獲得できたため、HiPerformer-w/o-classより良い結果となったわけである。
\textbf{Baseline Comparison.}
Results for the baseline and proposed methods are shown in \ref{tab:multiagent}.
The methods with ``min'' at the end of the model name are those that generate predictions probabilistically, and we report the best ADE and FDE among them.
For NLL, the average is reported. 
We generate 100 predictions for both datasets. 
% While this metric is common for evaluating generative methods~\citep{gupta2018social, mohamed2020social, li2021grin}, it is not fair to compare it with the results of nongenerative methods such as FQA and proposed methods; it would be fairer to compare using mean or median instead of the minimum value.
While this metric is common for evaluating generative methods~\citep{gupta2018social, mohamed2020social, li2021grin}, we cannot simply compare it with the results of nongenerative methods such as FQA and proposed methods; it would be better to compare using mean or median instead of the minimum value, and we bracket the min-model metrics in \ref{tab:multiagent}.
Therefore, for GRIN, the best-performing existing method, we report the average of the metrics for the generated predictions as ``GRIN-mean.'' 

In the Charged dataset, HiPerformer-w/o-class-prob has the best performance for all metrics and all the proposed methods outperform the GRIN-mean results.
% GRIN-min has the best performance for ADE and FDE, while HiPerformer-w/o-class-prob, HiPerformer-w/o-class-det, and HiPerformer-det exhibit close performance, and 
% Notably, all the proposed methods outperform the GRIN mean results.
Especially for NLL, HiPerformer-w/o-class-prob significantly exceeds baselines.
In this experiment, HiPerformer-w/o-class, which does not use class information, outperformed HiPerformer, which attempts to acquire hierarchical dependencies among series using the class structure.
This could be because the number of particles is small (5) and the relationships are simpler than those in the real-world dataset. Thus, HiPerformer-w/o-class could obtain richer information from other class particles using the raw data, rather than compressing each class information as in the 8-9 lines of \ref{alg:3dselfatt}.

For NBA, the proposed method also outperformed baselines in all indices. 
HiPerformer-det was best for ADE and FDE.
HiPerformer-w/o-class-prob and HiPerformer-prob performed similarly, both significantly outperforming baselines for NLL.

% HiPerformer-det and HiPerformer-prob performed particularly well, with HiPerformer-prob recording NLL less than 1/20 of GRIN. This may be because NBA players naturally play basketball with an awareness of other players' teams, and the classifications by team capture the actual dependency structure well.
% That is, HiPerformer successfully acquired that hierarchical dependencies, resulting in a better performance than HiPerformer-w/o-class.

% \textbf{Ablation Study.} 系列間の依存関係を獲得する$S_c, C$方向のSelf Attentionが有効であることを確認するための実験を行った。フルの提案モデルから$S_c, C$方向のSelf Attention Blockを削除し、$T$方向のみにSelf AttentionをとるAttTについての結果を\ref{tab:multiagent}の下部に示している。AttT-probは共分散を推定するが、AttT-detは平均のみを予測するモデルである。一貫してHiPerformerまたはHiPerformer-w/o-classがAttTの性能を上回っており、$S_c, C$方向のSelf Attentionが系列間の依存関係の獲得に有効に働いているといえる。
\textbf{Ablation Study.}
We conduct experiments to confirm the effectiveness of self-attention for $S_c, C$ dimension to acquire inter-series dependencies. 
The results for AttT, which removes the self-attention block for $S_c, C$ dimension from the full proposed model and takes self-attention for $T$ dimension only, are shown at the bottom of the \ref{tab:multiagent}.
AttT-prob estimates the mean and covariance, while AttT-det predicts only the mean.
Consistently, HiPerformer and HiPerformer-w/o-class outperforms AttT, indicating that self-attention for $S_c, C$ dimension is effective in obtaining dependencies among series.

\textbf{Remove Series.}
% NBAデータセットを使って、提案手法の系列数の変化に対する性能を調べた。
% オリジナルのテストデータからランダムに何人かの選手を取り除いて新たにテストデータを作り、これを用いてBaseline Comparisonで得られた学習済みのモデルを評価した。
% Baseline Comparisonで得られた学習済みのモデルに、何人かの選手を取り除いたデータをテストデータとして評価した。
% HiPerformer-w/o-class-probとHiPerformer-probを比較した結果を\ref{tab:remove}で示している。
% ADEでは、両者の差は小さい。
% 一方でFDEとNLLでは、HiPerformer-probの方が系列の削除に対して頑健であり、削除する人数を増やすほどHiPerformer-w/o-class-probより良い性能を示す傾向がある。
% 特に、NLLで両者の差は大きい。
% ここから、クラスインバランスが大きかったり、系列数が少なかったりする場合には、クラス間の関係を用いることで、性能の低下を軽減することができると言える。
We investigated the performance of the proposed method on changes in the number of series using the NBA dataset.
We created a new test dataset set by randomly removing some players from the original test dataset, and evaluated the trained models obtained in \textbf{Baseline Comparison}.
A comparison of HiPerformer-w/o-class-prob and HiPerformer-prob is shown in the \ref{tab:remove}.
For ADE, the difference between the two is small.
On the other hand, for FDE and NLL, HiPerformer-prob is more robust against removing series, and tends to perform better than HiPerformer-w/o-class-prob as more players are removed.
In particular, the difference between the two is large for NLL.
From this, we can say that when the class imbalance is large or the number of series is small, the performance degradation can be mitigated by using the relationship among classes.

\input{table/remove}

\subsubsection{Qualitative Evaluation}
\begin{figure*}[tb]
\vskip 0.2in
\begin{center}
\includegraphics[width=\columnwidth]{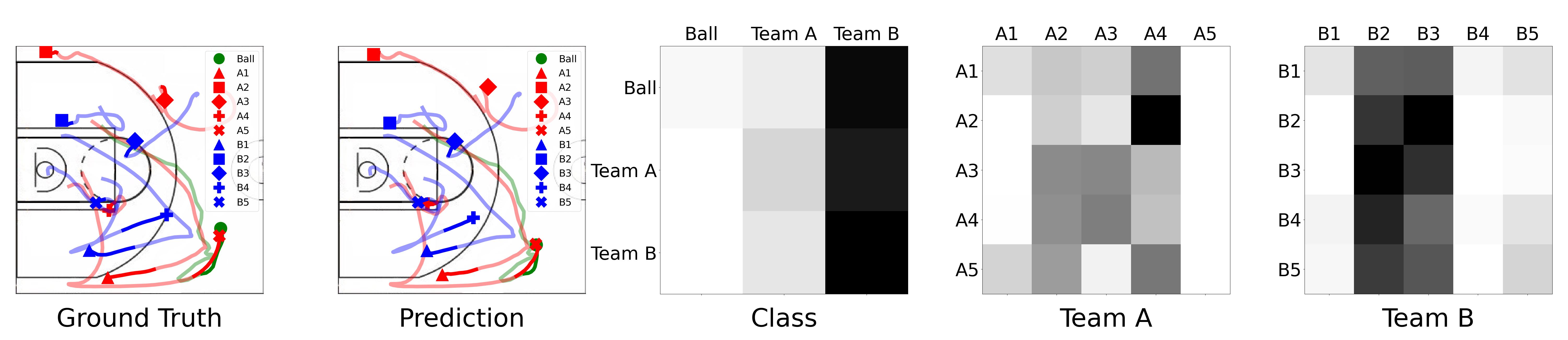}
\includegraphics[width=\columnwidth]{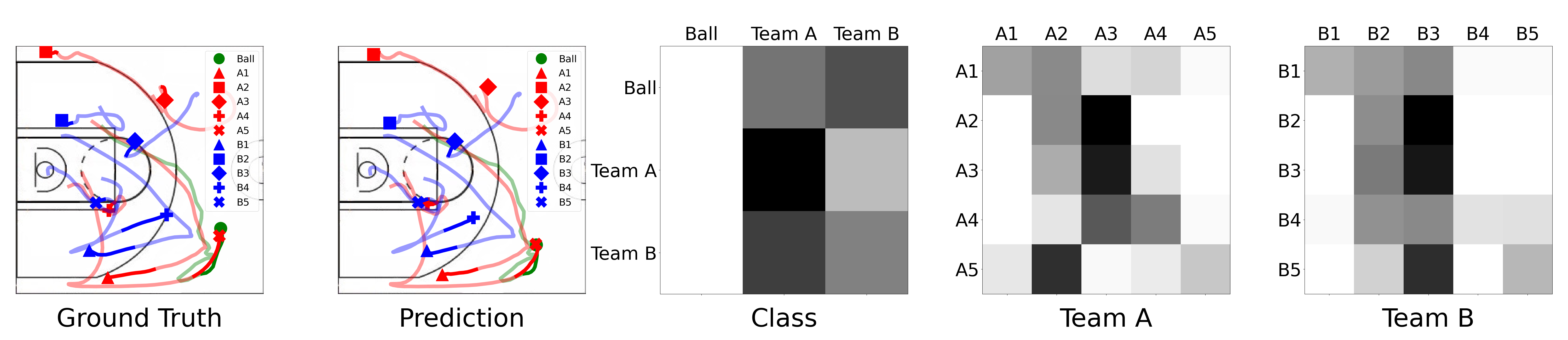}
\caption{
Trajectory prediction on NBA dataset. 
The left two figures are ground truth and predicted trajectories.
% 緑はボールを、赤はチームAの選手を、青はチームBの選手を表す。
% 薄い線は入力の40期間、濃い線は正解または予測の10期間の軌跡を表す。
% 左から3つ目の図は、クラス間のattention scoreを、4,5つ目は各チームの選手間のattention scoreを表す。
% 上段と下段は、同じ入力データにおける、違うヘッドのattention scoreである。
Green represents a ball, red represents a player from Team A, and blue represents a player from Team B.
The light line represents the 40-period trajectory of the input and the dark line represents the 10-period trajectory of the ground truth or prediction.
The third figure from the left represents the atttention score among the classes, and the fourth and fifth figures represent the atttention score among the players of each team.
Higher attention scores are in darker color.
The upper and lower rows show the attention scores of different heads on the same input data.
}
\label{fig:trajectory}
\end{center}
\vskip -0.2in
\end{figure*}

% NBAデータセットにおける軌跡の予測が\ref{fig:trajectory}で示されている。
% クラス間のattention scoreを見てみると、上段ではTeam Bに注目が集まり、下段ではTeam Aにも注目が集まるなど、多様な情報が取り出せているとわかる。
% 次に、上段のTeam Aの選手間のattention scoreのうちA5の選手に注目してみる。
% ここでA5はボールを持っている選手であり、敵のB4が近づいてきているため、見方の選手にボールをパスしようとしていると考えられる。
% A5が注目しているのはA1,A2,A4であり、これはA1,A2は敵選手が周りに少なく、A4はゴールに近い場所にいるからだと推測できる。
% 以上のような実世界の時系列の複雑な相互関係を、提案手法は捕捉できている。
The trajectory predictions in the NBA dataset are shown in \ref{fig:trajectory}.
Looking at the attention scores among classes, we can see that a wide variety of information can be retrieved, with the top row showing more attention to Team B and the bottom row showing more attention to Team A.
Next, take a look at the attention score of player A5 in the upper row.
Here, A5 is the player with the ball, and since the enemy B4 is approaching, he is probably trying to pass the ball to a player on his side.
Now, A5 is paying attention to A1, A2, and A4, and this may be because there are few enemy players around A1 and A2, and A4 is close to the goal.
The proposed method is able to capture the complex relationships among real-world time series as described above.

\subsection{Hierarchical Forecasting}

\input{table/hier_result.tex}

% \ref{sec:ex_multiagent}で使ったデータは、下から順に、bottom-level, class-level, root-levelの3階層からなるとみなせ、一番下のbottom-levelつまりagentについてのみ予測と評価を行った。それに対して今回は、上位階層の系列の特徴量が下位階層の系列の特徴量の和になっているような、4階層以上のデータで実験を行う。さらに、bottom-levelの系列だけではなく、各レベルの集約系列についても予測と評価を行う。提案手法においては、まずいずれかの階層をclass-levelに決めて、それに基づいてbottom-levelの系列をクラス分けし、\ref{sec:featureextractor}のFeature Extractorにより各bottom-level系列の特徴量を得る。集約系列の特徴量は、自分をrootとするsub-treeに含まれるbottom-level系列の特徴量の和とする。これで集約系列を含めた全系列の特徴量が得られるので、\ref{sec:distributionestimator}のDistribution Estimatorで分布を予測する。
The data used in \ref{sec:ex_multiagent} have three levels from bottom to top, namely, bottom, class, and root levels, and we make predictions and evaluations only for the bottom-level or agent. 
In contrast, at this time, we will conduct experiments on data with four or more levels, where the feature values of the upper-level series are the sum of those of the lower-level series.
Furthermore, we perform prediction and evaluation for the bottom-level series and the aggregate series at each level.

\subsubsection{Datasets}
% すべてのデータセットについて、$T$期間のデータが与えられて、$\tau$期間の予測をする場合、$1:T-2\tau$をtrain dataとし、$T-2\tau+1:T-\tau$においてvalidationを行い、$T-\tau+1:T$でテストを行う。これは、このデータセットを用いている既存研究に倣った設定である~\citep{rangapuram2021end, olivares2021probabilistic}。予測期間$\tau$についても同じく先行研究に倣う。\\
For all datasets, given data for $T$ periods, to make predictions for $\tau$ periods, we use $1:T-2\tau$ data in training, perform validation at $T-2\tau+1:T-\tau$, and test at $T-\tau+1:T$. This is a setting following existing studies using this dataset~\citep{rangapuram2021end, olivares2021probabilistic}, and we also adopt the same forecasting period $\tau$.

\textbf{Labour}~\citep{labour} is monthly Australian employment data from Feb. 1978 to Dec. 2020. We construct a 57-series hierarchy using categories such as region, gender, and employment status, as in ~\citep{rangapuram2021end}. 
In this data, $T$ is 514 and $\tau$ is 8.
% 4階層からなるデータで、提案手法においては上から2階層目のレベルでbottom-level系列をクラス分けした。\\
It has four levels, and we classify the bottom-level series at the second level from the top in the proposed method.

\textbf{Traffic}~\citep{cuturi2011fast, duagraff} records the occupancy of 963 car lanes in the San Francisco Bay Area freeways from January 2008 to March 2009. We obtain daily observations for one year and generate a 207-series hierarchy using the same aggregation method as in previous hierarchical forecasting research~\citep{ben2019regularized, rangapuram2021end}.
In these data, $T$ is 366 and $\tau$ is 1.
It has four levels, and we classify the bottom-level series at the second level from the top in the proposed method. 

\textbf{Wiki}\footnote{https://www.kaggle.com/c/web-traffic-time-series-forecasting/data} consists of daily views of 145K Wikipedia articles from July 2015 to December 2016. We follow previous studies~\citep{ben2019regularized, rangapuram2021end, olivares2021probabilistic} to select 150 bottom series and generate a 199-series hierarchy.
In these data, $T$ is 366 and $\tau$ is 1.
It has five levels, and we classify the bottom-level series at the fourth level from the top in the proposed method. 

\subsubsection{Evaluation Metrics}
% \ref{sec:multieval}と同じように、点推定だけではなく分布も評価するために、Root Mean Squared Error (RMSE)とNegative Log Likelihood (NLL)を採用する。どちらの評価指標についても、レベルごとに平均を計算する。
As with \ref{sec:multieval}, to evaluate the accuracy of probabilistic forecasts, we evaluate the likelihood of the forecast distribution with NLL and the results of point forecasts with RMSE. 
For both evaluation metrics, we calculate averages for each aggregation level.

\subsubsection{Methods} \label{sec:multiobjectmethods}
% このタスクにおいて成功している以下の3つの最近の手法をbaselineとして採用する。特に重要なポイントに対して、\ref{tab:methodscompare}で、baselinesと提案手法を比較している。\\
We adopt as baselines the following three recent methods that have been successful in this task.

% \textbf{DeepVAR}~\citep{salinas2019high}は、Vector Autoregressive ModelをRNNによって深層学習に拡張したものであり、multivariate time seriesが従う分布のパラメータを求めることができる確率的な予測手法である。階層的な一貫性は保証されない。\\
% \textbf{DeepVAR+}~\citep{rangapuram2021end}は、DeepVARの予測結果に、和に関する階層的な一貫性を持たせたものである。\\
% \textbf{HierE2E}~\citep{rangapuram2021end}は、まず対象系列集合の従う分布を推定し、そこからサンプリングされた予測を和に関して階層的に一貫するようにプロジェクションする手法である。知る限りこのタスクにおけるstate-of-the-artである。
\textbf{DeepVAR} ~\citep{salinas2019high} extends the VAM to deep learning with RNN, a probabilistic forecasting method that can determine the parameters of the distribution that a multivariate time series follows. Hierarchical consistency is not guaranteed.

\textbf{DeepVAR+} ~\citep{rangapuram2021end} adds hierarchical consistency to DeepVAR's prediction with respect to sums. 

\textbf{HierE2E} ~\citep{rangapuram2021end} first estimates the distribution followed by the target set of series and then projects the predictions sampled from it to be hierarchically consistent with respect to the sums. To the best of our knowledge, this is the state-of-the-art method for this task.

Next we explain about our proposed model HiPerformer.

\textbf{HiPerformer.}
We use full proposed model HiPerformer, which we call HiPerformer-prob in \ref{sec:ex_multiagent}.
We first select one of the aggregation levels as the class level and classify the bottom-level series based on it.
Subsequently, we obtain the feature values of each bottom-level series by feature extractor in \ref{sec:featureextractor}.
Here, the feature values of the aggregate series are the sum of the feature values of the bottom-level series in the sub-tree with itself as the root. 
Currently, we have the feature values of all the series including the aggregate series and predict the distribution using the distribution estimator in \ref{sec:distributionestimator}.
We use ISAB described in \ref{sec:permutation} to reduce computational cost when taking self-attention for $S_c, C$ dimension, and the dimension of inducing points ~\citep{lee2019} is set to 20.

A comparison of baselines with the proposed method is shown in \ref{tab:methodscompare}.

\textbf{Implementation Details.}
We use the same parameters in \ref{sec:multiobjectmethods} for our model, except for batch size.
We set the batch size 8 for labour, 4 for traffic, and 4 for wiki.
We train baseline models with parameters used in ~\citep{rangapuram2021end}.

% \subsubsection{Quantitative Evaluation}
\subsubsection{Evaluation}
% 結果を\ref{tab:hier}に示す。
% 見やすさのために、ベースラインの結果は最も良いもののみ載せている。
% Trafficのlevel 4のNLLを除いたすべてのデータセットと評価指標において、提案モデルがベースラインを上回った。
% 特にRMSEに関する性能の向上が顕著であり、和に関する階層的な一貫性を提案手法がよく捉えられていることがうかがえる。
% NLLについてもベースラインを大幅に上回る場合も多く、提案手法は、多階層なデータにおいても、その階層的な依存構造を考慮した多様な同時分布を出力できているとわかる。
The results are shown in the \ref{tab:hier}.
% The proposed model outperforms the baseline on all datasets and metrics except for Traffic's level 4 NLL.
The proposed model outperforms the baseline on all datasets and metrics except for the bottom-level NLL of Traffic and Wiki.
Particularly, the improvement in performance on RMSE is remarkable, indicating that the proposed method captures the hierarchical consistency for sums well.
The NLL is often markedly improved compared to the baseline, demonstrating the proposed method's capacity to generate diverse joint distributions while preserving the hierarchical dependency structure, even for multilevel data.

% \subsubsection{Quantitative Evaluation}
% to be written

%% file: table/methods_compare.tex
\begin{table*}[t]
% \caption {Baselineと提案手法の比較。確率的な予測が可能かどうか、系列間の階層的な依存構造を考慮できるか、系列間の依存構造を考慮できるか、系列の入退場に対応可能かどうか、階層的に順序共変かどうか、順序共変かどう、クラス情報を必要としないかどうか、で比較した。$\triangle$がついているのは、DeepVAR+とHierE2Eについては和に関してのみの階層的な依存構造を推定するため、HPEについては完全な順序共変性は保証されないが階層的に順序共変であるため、である。\todo{no class information neededは書き方大丈夫か？否定の疑問文の答え方的な危うさを感じる。}}
\caption{Comparison of baselines and the proposed method. The $\triangle$ is added because DeepVAR+ and HierE2E estimate hierarchical dependency structures only with respect to sums, and HiPerformer is hierarchically permutation-equivariant, although full permutation-equivariance is not guaranteed.}
\label {tab:methodscompare}
\vskip 0.1in
\begin{center}
\begin{scriptsize}
% \begin{sc}

\begin{tabular}{ccccccccc}
\hline
Method                                 & FQA          & NRI          & Social-GAN   & DeepVAR      & DeepVAR+     & HierE2E      & \textbf{\begin{tabular}[c]{@{}c@{}}HiPerformer\\ -w/o-class (Ours)\end{tabular}} & \textbf{\begin{tabular}[c]{@{}c@{}}HiPerformer\\ (Ours)\end{tabular}} \\ \hline
Probabilistic                          & $\times$     & $\checkmark$ & $\checkmark$ & $\checkmark$ & $\checkmark$ & $\checkmark$ & $\checkmark$                                                                     & $\checkmark$                                                          \\
Inter-Series Hierarchical Dependencies & $\times$     & $\times$     & $\times$     & $\times$     & $\triangle$  & $\triangle$  & $\times$                                                                         & $\checkmark$                                                          \\
Inter-Series Dependencies              & $\checkmark$ & $\checkmark$ & $\checkmark$ & $\checkmark$ & $\checkmark$ & $\checkmark$ & $\checkmark$                                                                     & $\checkmark$                                                          \\
Entry and Exit of Series               & $\times$     & $\times$     & $\checkmark$ & $\times$     & $\times$     & $\times$     & $\checkmark$                                                                     & $\checkmark$                                                          \\
Hierarchically-Permutation-Equivariant & $\times$     & $\times$     & $\times$     & $\times$     & $\times$     & $\times$     & $\times$                                                                         & $\checkmark$                                                          \\
Permutation-Equivariant                & $\times$     & $\times$     & $\checkmark$ & $\times$     & $\times$     & $\times$     & $\checkmark$                                                                     & $\triangle$                                                           \\
No Class Information Needed            & $\checkmark$ & $\checkmark$ & $\checkmark$ & $\checkmark$ & $\times$     & $\times$     & $\checkmark$                                                                     & $\times$                                                              \\ \hline
\end{tabular}

% \end{sc}
\end{scriptsize}
\end{center}
\vskip -0.1in
\end{table*}

%% file: table/multiagent_result.tex
\begin{table*}[t]
\caption {Performance comparison in multi-agent trajectory prediction. Non-generative methods have $\dagger$ after their names.　For GRIN and the proposed methods, we report the mean and standard deviation over five runs. The best results are in \textbf{bold}.}
\label {tab:multiagent}
\vskip 0.1in
\begin{center}
\begin{small}
% \begin{sc}

\begin{tabular}{c|ccc|ccc}
\hline
Dataset                                                                               & \multicolumn{3}{c|}{Charged}                                                        & \multicolumn{3}{c}{NBA}                                                            \\ \hline
Model                                                                                 & ADE                    & FDE                    & NLL                               & ADE                    & FDE                    & NLL                              \\ \hline
FQA$\dagger$                                                                          & 0.82                   & 1.76                   & -                                 & 2.42                   & 4.81                   & -                                \\
NRI-min                                                                               & (0.63)                 & (1.30)                 & -                                 & (2.10)                 & (4.56)                 & 586.9                            \\
Social-GAN-min                                                                        & (0.66)                 & (1.25)                 & -                                 & (1.88)                 & (3.64)                 & -                                \\
GRIN-min                                                                              & (0.53$\pm$0.01)        & (1.10$\pm$0.02)        & \multirow{2}{*}{237.86$\pm$30.51} & (1.73$\pm$0.03)        & (3.67$\pm$0.08)        & \multirow{2}{*}{509.27$\pm$3.23} \\
GRIN-mean                                                                             & 0.69$\pm$0.06          & 1.45$\pm$0.12          &                                   & 1.98$\pm$0.03          & 4.26$\pm$0.08          &                                  \\ \hline
\begin{tabular}[c]{@{}c@{}}HiPerformer-w/o-class\\ -det (Ours)$\dagger$\end{tabular}  & 0.51$\pm$0.01          & 1.14$\pm$0.02          & -                                 & 1.65$\pm$0.00          & 3.64$\pm$0.00          & -                                \\
HiPerformer-det (Ours)$\dagger$                                                       & 0.58$\pm$0.00          & 1.22$\pm$0.02          & -                                 & \textbf{1.64$\pm$0.01} & \textbf{3.60$\pm$0.02} & -                                \\
\begin{tabular}[c]{@{}c@{}}HiPerformer-w/o-class\\ -prob (Ours)$\dagger$\end{tabular} & \textbf{0.44$\pm$0.00} & \textbf{1.05$\pm$0.00} & \textbf{-11.70$\pm$1.57}          & \textbf{1.64$\pm$0.01} & 3.69$\pm$0.02          & \textbf{20.96$\pm$0.20}          \\
HiPerformer-prob (Ours)$\dagger$                                                      & 0.52$\pm$0.01          & 1.18$\pm$0.01          & 2.9$\pm$3.48                      & 1.65$\pm$0.00          & 3.68$\pm$0.01          & 21.14$\pm$0.26                   \\ \hline
AttT-det$\dagger$                                                                     & 0.63$\pm$0.01          & 1.37$\pm$0.02          & -                                 & 1.83$\pm$0.00          & 4.06$\pm$0.00          & -                                \\
AttT-prob$\dagger$                                                                    & 0.61$\pm$0.01          & 1.37$\pm$0.02          & 13.85$\pm$1.03                    & 1.92$\pm$0.00          & 4.25$\pm$0.00          & 26.58$\pm$0.14                   \\ \hline
\end{tabular}

% \end{sc}
\end{small}
\end{center}
\vskip -0.1in
\end{table*}

%% file: table/remove.tex
\begin{table*}[t]
\caption {
% NBAデータにおいて、A, Bチームからそれぞれ0-4人削除してテストしたときのHiPerformer-w/o-class-probとHiPerformer-probの性能の比較。
% 上の表から順にADE, FDE, NLLを表している。
% 学習は全選手のデータで行い、テスト時にランダムで選手を削除した。
% 各マスの左側がHiPerformer-w/o-class-probで、右側がHiPerformer-probの結果である。
Comparison of HiPerformer-w/o-class-prob and HiPerformer-prob performance in NBA dataset when tested with 0-4 players removed from team A and B, respectively.
In order from the table above, ADE, FDE, and NLL are represented.
Training was performed on all players' data, and players were randomly removed during the test phase.
The left side of each square is the result of HiPerformer-w/o-class-prob and the right side is the result of HiPerformer-prob.
We report the mean and standard deviation over five runs. 
The better results are in \textbf{bold}.
% Performance comparison in multi-agent trajectory prediction. Non-generative methods have $\dagger$ after their names.　For GRIN and the proposed methods, we report the mean and standard deviation over five runs. The best results are in \textbf{bold}.
}
\label {tab:remove}
\vskip 0.1in
\begin{center}
\begin{scriptsize}
% \begin{sc}

\begin{tabular}{cccccc}
\hline
\multicolumn{1}{c|}{\diagbox{A}{ADE}{B}} & \multicolumn{1}{c|}{0}                                        & \multicolumn{1}{c|}{1}                                        & \multicolumn{1}{c|}{2}                                        & \multicolumn{1}{c|}{3}                                        & 4                                        \\ \hline
\multicolumn{1}{c|}{0}                   & \multicolumn{1}{c|}{(1.64$\pm$0.01 - 1.65$\pm$0.00)}          & \multicolumn{1}{c|}{1.69$\pm$0.01 - 1.69$\pm$0.01}            & \multicolumn{1}{c|}{1.73$\pm$0.01 - 1.73$\pm$0.01}            & \multicolumn{1}{c|}{1.78$\pm$0.01 - 1.78$\pm$0.01}            & 1.85$\pm$0.01 - \textbf{1.84$\pm$0.01}   \\ \hline
\multicolumn{1}{c|}{1}                   & \multicolumn{1}{c|}{\textbf{1.67$\pm$0.01} - 1.68$\pm$0.01}   & \multicolumn{1}{c|}{\textbf{1.71$\pm$0.01} - 1.72$\pm$0.01}   & \multicolumn{1}{c|}{\textbf{1.76$\pm$0.01} - 1.77$\pm$0.01}   & \multicolumn{1}{c|}{1.83$\pm$0.01 - 1.83$\pm$0.01}            & 1.92$\pm$0.01 - \textbf{1.91$\pm$0.01}   \\ \hline
\multicolumn{1}{c|}{2}                   & \multicolumn{1}{c|}{\textbf{1.69$\pm$0.01} - 1.71$\pm$0.01}   & \multicolumn{1}{c|}{\textbf{1.74$\pm$0.01} - 1.76$\pm$0.01}   & \multicolumn{1}{c|}{\textbf{1.81$\pm$0.01} - 1.82$\pm$0.01}   & \multicolumn{1}{c|}{1.89$\pm$0.01 - 1.89$\pm$0.01}            & 2.01$\pm$0.01 - 2.01$\pm$0.01            \\ \hline
\multicolumn{1}{c|}{3}                   & \multicolumn{1}{c|}{\textbf{1.73$\pm$0.01} - 1.74$\pm$0.01}   & \multicolumn{1}{c|}{\textbf{1.79$\pm$0.01} - 1.80$\pm$0.01}   & \multicolumn{1}{c|}{\textbf{1.87$\pm$0.01} - 1.88$\pm$0.01}   & \multicolumn{1}{c|}{\textbf{1.98$\pm$0.01} - 1.99$\pm$0.01}   & \textbf{2.15$\pm$0.01} - 2.16$\pm$0.01   \\ \hline
\multicolumn{1}{c|}{4}                   & \multicolumn{1}{c|}{\textbf{1.76$\pm$0.01} - 1.79$\pm$0.01}   & \multicolumn{1}{c|}{\textbf{1.84$\pm$0.01} - 1.87$\pm$0.01}   & \multicolumn{1}{c|}{\textbf{1.95$\pm$0.01} - 1.97$\pm$0.01}   & \multicolumn{1}{c|}{\textbf{2.11$\pm$0.01} - 2.13$\pm$0.01}   & \textbf{2.38$\pm$0.01} - 2.40$\pm$0.01   \\ \hline
                                         &                                                               &                                                               &                                                               &                                                               &                                          \\ \hline
\multicolumn{1}{c|}{\diagbox{A}{FDE}{B}} & \multicolumn{1}{c|}{0}                                        & \multicolumn{1}{c|}{1}                                        & \multicolumn{1}{c|}{2}                                        & \multicolumn{1}{c|}{3}                                        & 4                                        \\ \hline
\multicolumn{1}{c|}{0}                   & \multicolumn{1}{c|}{(3.69$\pm$0.02 - 3.68$\pm$0.01)}          & \multicolumn{1}{c|}{3.78$\pm$0.02 - \textbf{3.76$\pm$0.01}}   & \multicolumn{1}{c|}{3.87$\pm$0.02 - \textbf{3.84$\pm$0.01}}   & \multicolumn{1}{c|}{3.98$\pm$0.02 - \textbf{3.94$\pm$0.01}}   & 4.13$\pm$0.02 - \textbf{4.07$\pm$0.01}   \\ \hline
\multicolumn{1}{c|}{1}                   & \multicolumn{1}{c|}{3.73$\pm$0.02 - \textbf{3.72$\pm$0.01}}   & \multicolumn{1}{c|}{3.82$\pm$0.02 - \textbf{3.80$\pm$0.01}}   & \multicolumn{1}{c|}{3.93$\pm$0.02 - \textbf{3.90$\pm$0.01}}   & \multicolumn{1}{c|}{4.06$\pm$0.02 - \textbf{4.02$\pm$0.01}}   & 4.25$\pm$0.02 - \textbf{4.19$\pm$0.01}   \\ \hline
\multicolumn{1}{c|}{2}                   & \multicolumn{1}{c|}{3.76$\pm$0.02 - 3.76$\pm$0.01}            & \multicolumn{1}{c|}{3.86$\pm$0.02 - \textbf{3.85$\pm$0.01}}   & \multicolumn{1}{c|}{3.99$\pm$0.02 - \textbf{3.97$\pm$0.01}}   & \multicolumn{1}{c|}{4.16$\pm$0.02 - \textbf{4.13$\pm$0.01}}   & 4.41$\pm$0.02 - \textbf{4.35$\pm$0.01}   \\ \hline
\multicolumn{1}{c|}{3}                   & \multicolumn{1}{c|}{3.81$\pm$0.02 - 3.81$\pm$0.01}            & \multicolumn{1}{c|}{3.93$\pm$0.02 - 3.93$\pm$0.01}            & \multicolumn{1}{c|}{4.09$\pm$0.02 - \textbf{4.08$\pm$0.01}}   & \multicolumn{1}{c|}{4.31$\pm$0.02 - \textbf{4.29$\pm$0.01}}   & 4.65$\pm$0.02 - \textbf{4.61$\pm$0.01}   \\ \hline
\multicolumn{1}{c|}{4}                   & \multicolumn{1}{c|}{\textbf{3.86$\pm$0.02} - 3.87$\pm$0.01}   & \multicolumn{1}{c|}{\textbf{4.01$\pm$0.02} - 4.02$\pm$0.01}   & \multicolumn{1}{c|}{4.22$\pm$0.02 - 4.22$\pm$0.02}            & \multicolumn{1}{c|}{4.53$\pm$0.02 - \textbf{4.52$\pm$0.02}}   & 5.06$\pm$0.02 - \textbf{5.02$\pm$0.02}   \\ \hline
                                         &                                                               &                                                               &                                                               &                                                               &                                          \\ \hline
\multicolumn{1}{c|}{\diagbox{A}{NLL}{B}} & \multicolumn{1}{c|}{0}                                        & \multicolumn{1}{c|}{1}                                        & \multicolumn{1}{c|}{2}                                        & \multicolumn{1}{c|}{3}                                        & 4                                        \\ \hline
\multicolumn{1}{c|}{0}                   & \multicolumn{1}{c|}{(20.96$\pm$0.20 - 21.14$\pm$0.26)}        & \multicolumn{1}{c|}{21.43$\pm$0.29 - \textbf{21.40$\pm$0.23}} & \multicolumn{1}{c|}{21.89$\pm$0.32 - \textbf{21.79$\pm$0.24}} & \multicolumn{1}{c|}{22.45$\pm$0.35 - \textbf{22.26$\pm$0.24}} & 22.97$\pm$0.39 - \textbf{22.67$\pm$0.24} \\ \hline
\multicolumn{1}{c|}{1}                   & \multicolumn{1}{c|}{\textbf{21.39$\pm$0.30} - 21.42$\pm$0.24} & \multicolumn{1}{c|}{21.74$\pm$0.31 - \textbf{21.69$\pm$0.23}} & \multicolumn{1}{c|}{22.29$\pm$0.35 - \textbf{22.16$\pm$0.24}} & \multicolumn{1}{c|}{22.99$\pm$0.40 - \textbf{22.75$\pm$0.25}} & 23.69$\pm$0.45 - \textbf{23.31$\pm$0.25} \\ \hline
\multicolumn{1}{c|}{2}                   & \multicolumn{1}{c|}{21.68$\pm$0.33 - 21.68$\pm$0.22}          & \multicolumn{1}{c|}{22.10$\pm$0.34 - \textbf{22.01$\pm$0.22}} & \multicolumn{1}{c|}{22.78$\pm$0.39 - \textbf{22.60$\pm$0.23}} & \multicolumn{1}{c|}{23.68$\pm$0.45 - \textbf{23.36$\pm$0.24}} & 24.65$\pm$0.52 - \textbf{24.15$\pm$0.23} \\ \hline
\multicolumn{1}{c|}{3}                   & \multicolumn{1}{c|}{22.04$\pm$0.37 - \textbf{22.01$\pm$0.22}} & \multicolumn{1}{c|}{22.57$\pm$0.39 - \textbf{22.44$\pm$0.21}} & \multicolumn{1}{c|}{23.45$\pm$0.45 - \textbf{23.19$\pm$0.22}} & \multicolumn{1}{c|}{24.66$\pm$0.54 - \textbf{24.22$\pm$0.23}} & 26.12$\pm$0.65 - \textbf{25.44$\pm$0.22} \\ \hline
\multicolumn{1}{c|}{4}                   & \multicolumn{1}{c|}{22.51$\pm$0.41 - \textbf{22.43$\pm$0.20}} & \multicolumn{1}{c|}{23.21$\pm$0.44 - \textbf{23.00$\pm$0.19}} & \multicolumn{1}{c|}{24.39$\pm$0.53 - \textbf{24.02$\pm$0.19}} & \multicolumn{1}{c|}{26.13$\pm$0.65 - \textbf{25.51$\pm$0.21}} & 28.57$\pm$0.84 - \textbf{27.56$\pm$0.17} \\ \hline
\end{tabular}

% \end{sc}
\end{scriptsize}
\end{center}
\vskip -0.1in
\end{table*}

%% file: table/hier_result.tex
\begin{table*}[t]
\caption {RMSE and NLL at each aggregation level. We report the mean and standard deviation for proposed method and best of baselines over 5 runs. The best results are in \textbf{bold}.}
\label {tab:hier}
\vskip 0.1in
\begin{center}
\begin{scriptsize}
% \begin{sc}

\begin{tabular}{cccccc|cccc}
\hline
\multicolumn{2}{c}{Metrics}                                                & \multicolumn{4}{c|}{RMSE}                                                                                                                                                        & \multicolumn{4}{c}{NLL}                                                                                                                                                                \\ \hline
\multicolumn{1}{c|}{Dataset}                  & \multicolumn{1}{c|}{Level} & \begin{tabular}[c]{@{}c@{}}DeepVAR\\ (not coherent)\end{tabular} & DeepVAR+          & HierE2E           & \textbf{\begin{tabular}[c]{@{}c@{}}HiPerformer\\ (Ours)\end{tabular}} & \begin{tabular}[c]{@{}c@{}}DeepVAR\\ (not coherent)\end{tabular} & DeepVAR+         & HierE2E                  & \textbf{\begin{tabular}[c]{@{}c@{}}HiPerformer\\ (Ours)\end{tabular}} \\ \hline
\multicolumn{1}{c|}{\multirow{4}{*}{Labour}}  & \multicolumn{1}{c|}{1}     & 561.7 $\pm$ 59.2                                                 & 629.0 $\pm$ 117.0 & 656.3 $\pm$ 151.5 & \textbf{4.58$\pm$2.05}                                                & 13.11 $\pm$ 1.19                                                 & 12.14 $\pm$ 3.12 & 14.64 $\pm$ 3.54         & \textbf{4.33$\pm$0.41}                                                \\
\multicolumn{1}{c|}{}                         & \multicolumn{1}{c|}{2}     & 104.5 $\pm$ 13.5                                                 & 112.9 $\pm$ 15.3  & 112.5 $\pm$ 18.3  & \textbf{2.64$\pm$0.89}                                                & 7.44 $\pm$ 0.95                                                  & 7.79 $\pm$ 1.79  & 7.85 $\pm$ 1.23          & \textbf{3.88$\pm$0.55}                                                \\
\multicolumn{1}{c|}{}                         & \multicolumn{1}{c|}{3}     & 55.8 $\pm$ 6.5                                                   & 57.0 $\pm$ 7.7    & 57.4 $\pm$ 8.9    & \textbf{1.92$\pm$0.63}                                                & 6.49 $\pm$ 0.81                                                  & 6.25 $\pm$ 1.19  & 6.52 $\pm$ 0.89          & \textbf{3.58$\pm$0.37}                                                \\
\multicolumn{1}{c|}{}                         & \multicolumn{1}{c|}{4}     & 31.4 $\pm$ 3.5                                                   & 31.9 $\pm$ 3.4    & 32.62 $\pm$ 4.55  & \textbf{1.57$\pm$0.45}                                                & 4.79 $\pm$ 0.36                                                  & 4.81 $\pm$ 0.56  & 5.14 $\pm$ 0.49          & \textbf{4.28$\pm$0.32}                                                \\ \hline
\multicolumn{1}{c|}{\multirow{4}{*}{Traffic}} & \multicolumn{1}{c|}{1}     & 41.90 $\pm$ 8.09                                                 & 22.40 $\pm$ 8.44  & 22.59 $\pm$ 15.71 & \textbf{3.10$\pm$2.26}                                                & 5.19 $\pm$ 0.15                                                  & 4.65 $\pm$ 0.18  & 4.77 $\pm$ 0.36          & \textbf{3.96$\pm$0.84}                                                \\
\multicolumn{1}{c|}{}                         & \multicolumn{1}{c|}{2}     & 24.79 $\pm$ 8.16                                                 & 14.66 $\pm$ 1.98  & 15.13 $\pm$ 5.70  & \textbf{2.21$\pm$1.59}                                                & 4.65 $\pm$ 0.31                                                  & 4.20 $\pm$ 0.08  & 4.19 $\pm$ 0.25          & \textbf{3.50$\pm$0.67}                                                \\
\multicolumn{1}{c|}{}                         & \multicolumn{1}{c|}{3}     & 17.35 $\pm$ 3.98                                                 & 10.70 $\pm$ 3.89  & 9.82 $\pm$ 2.29   & \textbf{1.56$\pm$1.12}                                                & 4.28 $\pm$ 0.41                                                  & 3.92 $\pm$ 0.22  & 3.71 $\pm$ 0.18          & \textbf{3.43$\pm$0.77}                                                \\
\multicolumn{1}{c|}{}                         & \multicolumn{1}{c|}{4}     & 1.46 $\pm$ 0.08                                                  & 1.44 $\pm$ 0.05   & 1.47 $\pm$ 0.03   & \textbf{0.29$\pm$0.12}                                                & 1.65 $\pm$ 0.14                                                  & 1.57 $\pm$ 0.07  & \textbf{1.44 $\pm$ 0.12} & 3.99$\pm$1.35                                                         \\ \hline
\multicolumn{1}{c|}{\multirow{5}{*}{Wiki}}    & \multicolumn{1}{c|}{1}     & 11088 $\pm$ 6717                                                 & 11721 $\pm$ 7688  & 4125 $\pm$ 2483   & \textbf{204.73$\pm$175.54}                                            & 11.05 $\pm$ 0.44                                                 & 11.50 $\pm$ 0.36 & 9.88 $\pm$ 0.35          & \textbf{8.84$\pm$1.48}                                                \\
\multicolumn{1}{c|}{}                         & \multicolumn{1}{c|}{2}     & 3107 $\pm$ 941                                                   & 4188 $\pm$ 1218   & 3855 $\pm$ 340    & \textbf{158.68$\pm$91.94}                                             & 9.57 $\pm$ 0.20                                                  & 10.03 $\pm$ 0.24 & 12.66 $\pm$ 1.34         & \textbf{8.13$\pm$0.45}                                                \\
\multicolumn{1}{c|}{}                         & \multicolumn{1}{c|}{3}     & 2773 $\pm$ 60                                                    & 3082 $\pm$ 335    & 2851 $\pm$ 187    & \textbf{142.58$\pm$36.72}                                             & 9.69 $\pm$ 0.17                                                  & 9.28 $\pm$ 0.12  & 12.85 $\pm$ 0.92         & \textbf{8.12$\pm$0.43}                                                \\
\multicolumn{1}{c|}{}                         & \multicolumn{1}{c|}{4}     & 2737 $\pm$ 178                                                   & 2943 $\pm$ 264    & 2769 $\pm$ 151    & \textbf{145.90$\pm$37.89}                                              & 9.45 $\pm$ 0.46                                                  & 9.21 $\pm$ 0.12  & 15.56 $\pm$ 1.68         & \textbf{8.11$\pm$0.43}                                                \\
\multicolumn{1}{c|}{}                         & \multicolumn{1}{c|}{5}     & 931 $\pm$ 22                                                     & 978 $\pm$ 65      & 932 $\pm$ 26      & \textbf{96.60$\pm$22.26}                                               & \textbf{8.02 $\pm$ 0.20}                                         & 8.17 $\pm$ 0.43  & 9.61 $\pm$ 0.84          & 8.26$\pm$0.24                                                         \\ \hline
\end{tabular}

% \end{sc}
\end{scriptsize}
\end{center}
\vskip -0.1in
\end{table*}

%% file: section/5_conclusion.tex
\section{Conclusion}
% 私たちは、系列の新規参入や退場が起こりうる時系列の集合に対して適用可能な、2つの予測モデルを提案した。一つ目は、系列間の階層的な依存関係を捕捉可能なモデルである。その中で階層的な順序変換および階層的な順序共変性という新しい概念を定義し、提案モデルが階層的に順序共変であることを示した。二つ目は、クラス情報を用いずに系列間の関係を捉える、系列方向に順序共変なモデルである。どちらのモデルも時変な共分散を含む多様な同時分布を出力することができる。これらに対して、2種類のタスクにおける5つのデータでの実験の定性的、定量的な評価を行い、提案モデルの有効性を確認するとともに、クラス分けされている時系列集合の予測において、その階層的な依存構造の活用が有効であることを示した。
We proposed two prediction models HiPerformer and HiPerformer-w/o-class that can be applied to a set of time series in which new entries and exits of series can occur. 
The first model is HiPerformer that can capture hierarchical dependencies among series.
To achieve this property, we defined the new concepts of hierarchical permutation and hierarchical permutation-equivariance and showed that HiPerformer is hierarchically permutation-equivariant.
The second model is HiPerformer-w/o-class that is permutation-equivariant and captures the relationships among series without using class information.
Both models can output various joint distributions, including time-varying covariances. 
These proposed models achieve state-of-the-art performance on synthetic and real-world datasets.
From this, we indicate that using its hierarchical dependency structure is effective in forecasting classified time series sets.

%% file: section/6_acknowledgements.tex
\\\\
\textbf{Acknowledgements}\\
This work was partially supported by JST AIP Acceleration Research JPMJCR20U3, Moonshot R\&D Grant Number JPMJPS2011, CREST Grant Number JPMJCR2015, JSPS KAKENHI Grant Number JP19H01115 and Basic Research Grant (Super AI) of Institute for AI and Beyond of the University of Tokyo.

%% file: main.bbl
\begin{thebibliography}{42}
\providecommand{\natexlab}[1]{#1}
\providecommand{\url}[1]{\texttt{#1}}
\expandafter\ifx\csname urlstyle\endcsname\relax
  \providecommand{\doi}[1]{doi: #1}\else
  \providecommand{\doi}{doi: \begingroup \urlstyle{rm}\Url}\fi

\bibitem[Sukcharoen and Leatham(2016)]{sukcharoen2016dependence}
Kunlapath Sukcharoen and David~J Leatham.
\newblock Dependence and extreme correlation among us industry sectors.
\newblock \emph{Studies in Economics and Finance}, 33\penalty0 (1):\penalty0
  26--49, 2016.

\bibitem[Zaheer et~al.(2017)Zaheer, Kottur, Ravanbakhsh, Poczos, Salakhutdinov,
  and Smola]{deepsets}
Manzil Zaheer, Satwik Kottur, Siamak Ravanbakhsh, Barnabas Poczos, Russ~R
  Salakhutdinov, and Alexander~J Smola.
\newblock Deep sets.
\newblock In \emph{Advances in Neural Information Processing Systems},
  volume~30, pages 3391--3401, 2017.

\bibitem[Cho et~al.(2014)Cho, van Merrienboer, G{\"{u}}l{\c{c}}ehre, Bougares,
  Schwenk, and Bengio]{DBLP:journals/corr/ChoMGBSB14}
Kyunghyun Cho, Bart van Merrienboer, {\c{C}}aglar G{\"{u}}l{\c{c}}ehre, Fethi
  Bougares, Holger Schwenk, and Yoshua Bengio.
\newblock Learning phrase representations using {RNN} encoder-decoder for
  statistical machine translation.
\newblock \emph{CoRR}, abs/1406.1078, 2014.
\newblock URL \url{http://arxiv.org/abs/1406.1078}.

\bibitem[Khandelwal et~al.(2018)Khandelwal, He, Qi, and
  Jurafsky]{khandelwal2018sharp}
Urvashi Khandelwal, He~He, Peng Qi, and Dan Jurafsky.
\newblock Sharp nearby, fuzzy far away: How neural language models use context.
\newblock \emph{arXiv preprint arXiv:1805.04623}, 2018.

\bibitem[Vaswani et~al.(2017)Vaswani, Shazeer, Parmar, Uszkoreit, Jones, Gomez,
  Kaiser, and Polosukhin]{vaswani2017}
Ashish Vaswani, Noam Shazeer, Niki Parmar, Jakob Uszkoreit, Llion Jones,
  Aidan~N Gomez, \L~ukasz Kaiser, and Illia Polosukhin.
\newblock Attention is all you need.
\newblock In \emph{Advances in Neural Information Processing Systems},
  volume~30, pages 5998--6008, 2017.

\bibitem[Li et~al.(2019)Li, Jin, Xuan, Zhou, Chen, Wang, and
  Yan]{li2019enhancing}
Shiyang Li, Xiaoyong Jin, Yao Xuan, Xiyou Zhou, Wenhu Chen, Yu-Xiang Wang, and
  Xifeng Yan.
\newblock Enhancing the locality and breaking the memory bottleneck of
  transformer on time series forecasting.
\newblock \emph{Advances in neural information processing systems}, 32, 2019.

\bibitem[Zhou et~al.(2021)Zhou, Zhang, Peng, Zhang, Li, Xiong, and
  Zhang]{zhou2021informer}
Haoyi Zhou, Shanghang Zhang, Jieqi Peng, Shuai Zhang, Jianxin Li, Hui Xiong,
  and Wancai Zhang.
\newblock Informer: Beyond efficient transformer for long sequence time-series
  forecasting.
\newblock In \emph{Proceedings of the AAAI Conference on Artificial
  Intelligence}, volume~35, pages 11106--11115, 2021.

\bibitem[Wu et~al.(2021)Wu, Xu, Wang, and Long]{wu2021autoformer}
Haixu Wu, Jiehui Xu, Jianmin Wang, and Mingsheng Long.
\newblock Autoformer: Decomposition transformers with auto-correlation for
  long-term series forecasting.
\newblock \emph{Advances in Neural Information Processing Systems},
  34:\penalty0 22419--22430, 2021.

\bibitem[Zhou et~al.(2022)Zhou, Ma, Wen, Wang, Sun, and Jin]{zhou2022fedformer}
Tian Zhou, Ziqing Ma, Qingsong Wen, Xue Wang, Liang Sun, and Rong Jin.
\newblock Fedformer: Frequency enhanced decomposed transformer for long-term
  series forecasting.
\newblock \emph{arXiv preprint arXiv:2201.12740}, 2022.

\bibitem[Liu et~al.(2021)Liu, Yu, Liao, Li, Lin, Liu, and
  Dustdar]{liu2021pyraformer}
Shizhan Liu, Hang Yu, Cong Liao, Jianguo Li, Weiyao Lin, Alex~X Liu, and
  Schahram Dustdar.
\newblock Pyraformer: Low-complexity pyramidal attention for long-range time
  series modeling and forecasting.
\newblock In \emph{International Conference on Learning Representations}, 2021.

\bibitem[Wu et~al.(2020)Wu, Xiao, Ding, Zhao, Wei, and
  Huang]{wu2020adversarial}
Sifan Wu, Xi~Xiao, Qianggang Ding, Peilin Zhao, Ying Wei, and Junzhou Huang.
\newblock Adversarial sparse transformer for time series forecasting.
\newblock \emph{Advances in neural information processing systems},
  33:\penalty0 17105--17115, 2020.

\bibitem[Goodfellow et~al.(2020)Goodfellow, Pouget-Abadie, Mirza, Xu,
  Warde-Farley, Ozair, Courville, and Bengio]{goodfellow2020generative}
Ian Goodfellow, Jean Pouget-Abadie, Mehdi Mirza, Bing Xu, David Warde-Farley,
  Sherjil Ozair, Aaron Courville, and Yoshua Bengio.
\newblock Generative adversarial networks.
\newblock \emph{Communications of the ACM}, 63\penalty0 (11):\penalty0
  139--144, 2020.

\bibitem[Scarselli et~al.(2008)Scarselli, Gori, Tsoi, Hagenbuchner, and
  Monfardini]{scarselli2008graph}
Franco Scarselli, Marco Gori, Ah~Chung Tsoi, Markus Hagenbuchner, and Gabriele
  Monfardini.
\newblock The graph neural network model.
\newblock \emph{IEEE transactions on neural networks}, 20\penalty0
  (1):\penalty0 61--80, 2008.

\bibitem[Bai et~al.(2020)Bai, Yao, Li, Wang, and Wang]{bai2020adaptive}
Lei Bai, Lina Yao, Can Li, Xianzhi Wang, and Can Wang.
\newblock Adaptive graph convolutional recurrent network for traffic
  forecasting.
\newblock \emph{Advances in neural information processing systems},
  33:\penalty0 17804--17815, 2020.

\bibitem[Defferrard et~al.(2016)Defferrard, Bresson, and
  Vandergheynst]{defferrard2016convolutional}
Micha{\"e}l Defferrard, Xavier Bresson, and Pierre Vandergheynst.
\newblock Convolutional neural networks on graphs with fast localized spectral
  filtering.
\newblock \emph{Advances in neural information processing systems}, 29, 2016.

\bibitem[Kipf and Welling(2016)]{kipf2016semi}
Thomas~N Kipf and Max Welling.
\newblock Semi-supervised classification with graph convolutional networks.
\newblock \emph{arXiv preprint arXiv:1609.02907}, 2016.

\bibitem[Zhao et~al.(2021)Zhao, Gao, Lan, Sun, Sapp, Varadarajan, Shen, Shen,
  Chai, Schmid, et~al.]{zhao2021tnt}
Hang Zhao, Jiyang Gao, Tian Lan, Chen Sun, Ben Sapp, Balakrishnan Varadarajan,
  Yue Shen, Yi~Shen, Yuning Chai, Cordelia Schmid, et~al.
\newblock Tnt: Target-driven trajectory prediction.
\newblock In \emph{Conference on Robot Learning}, pages 895--904. PMLR, 2021.

\bibitem[Salzmann et~al.(2020)Salzmann, Ivanovic, Chakravarty, and
  Pavone]{salzmann2020trajectron++}
Tim Salzmann, Boris Ivanovic, Punarjay Chakravarty, and Marco Pavone.
\newblock Trajectron++: Dynamically-feasible trajectory forecasting with
  heterogeneous data.
\newblock In \emph{European Conference on Computer Vision}, pages 683--700.
  Springer, 2020.

\bibitem[Mangalam et~al.(2020)Mangalam, An, Girase, and Malik]{2012.01526}
Karttikeya Mangalam, Yang An, Harshayu Girase, and Jitendra Malik.
\newblock {From Goals, Waypoints \& Paths To Long Term Human Trajectory
  Forecasting}, 2020.

\bibitem[Leon and Gavrilescu(2021)]{leon2021review}
Florin Leon and Marius Gavrilescu.
\newblock A review of tracking and trajectory prediction methods for autonomous
  driving.
\newblock \emph{Mathematics}, 9\penalty0 (6):\penalty0 660, 2021.

\bibitem[Kretzschmar et~al.(2014)Kretzschmar, Kuderer, and
  Burgard]{kretzschmar2014learning}
Henrik Kretzschmar, Markus Kuderer, and Wolfram Burgard.
\newblock Learning to predict trajectories of cooperatively navigating agents.
\newblock In \emph{2014 IEEE international conference on robotics and
  automation (ICRA)}, pages 4015--4020. IEEE, 2014.

\bibitem[Schmerling et~al.(2018)Schmerling, Leung, Vollprecht, and
  Pavone]{schmerling2018multimodal}
Edward Schmerling, Karen Leung, Wolf Vollprecht, and Marco Pavone.
\newblock Multimodal probabilistic model-based planning for human-robot
  interaction.
\newblock In \emph{2018 IEEE International Conference on Robotics and
  Automation (ICRA)}, pages 3399--3406. IEEE, 2018.

\bibitem[Felsen et~al.(2017)Felsen, Agrawal, and Malik]{felsen2017will}
Panna Felsen, Pulkit Agrawal, and Jitendra Malik.
\newblock What will happen next? forecasting player moves in sports videos.
\newblock In \emph{Proceedings of the IEEE international conference on computer
  vision}, pages 3342--3351, 2017.

\bibitem[Li et~al.(2021)Li, Yao, Wenliang, He, Xiao, Yan, Wipf, and
  Zhang]{li2021grin}
Longyuan Li, Jian Yao, Li~Wenliang, Tong He, Tianjun Xiao, Junchi Yan, David
  Wipf, and Zheng Zhang.
\newblock Grin: Generative relation and intention network for multi-agent
  trajectory prediction.
\newblock \emph{Advances in Neural Information Processing Systems},
  34:\penalty0 27107--27118, 2021.

\bibitem[Velickovic et~al.(2017)Velickovic, Cucurull, Casanova, Romero, Lio,
  and Bengio]{velickovic2017graph}
Petar Velickovic, Guillem Cucurull, Arantxa Casanova, Adriana Romero, Pietro
  Lio, and Yoshua Bengio.
\newblock Graph attention networks.
\newblock \emph{stat}, 1050:\penalty0 20, 2017.

\bibitem[Rangapuram et~al.(2021)Rangapuram, Werner, Benidis, Mercado, Gasthaus,
  and Januschowski]{rangapuram2021end}
Syama~Sundar Rangapuram, Lucien~D Werner, Konstantinos Benidis, Pedro Mercado,
  Jan Gasthaus, and Tim Januschowski.
\newblock End-to-end learning of coherent probabilistic forecasts for
  hierarchical time series.
\newblock In \emph{International Conference on Machine Learning}, pages
  8832--8843. PMLR, 2021.

\bibitem[Olivares et~al.(2021)Olivares, Meetei, Ma, Reddy, Cao, and
  Dicker]{olivares2021probabilistic}
Kin~G Olivares, O~Nganba Meetei, Ruijun Ma, Rohan Reddy, Mengfei Cao, and Lee
  Dicker.
\newblock Probabilistic hierarchical forecasting with deep poisson mixtures.
\newblock \emph{arXiv preprint arXiv:2110.13179}, 2021.

\bibitem[Lee et~al.(2019)Lee, Lee, Kim, Kosiorek, Choi, and Teh]{lee2019}
Juho Lee, Yoonho Lee, Jungtaek Kim, Adam Kosiorek, Seungjin Choi, and Yee~Whye
  Teh.
\newblock Set transformer: A framework for attention-based
  permutation-invariant neural networks.
\newblock In \emph{Proceedings of the 36th International Conference on Machine
  Learning}, volume~97, pages 3744--3753, 2019.

\bibitem[Kipf et~al.(2018)Kipf, Fetaya, Wang, Welling, and
  Zemel]{kipf2018neural}
Thomas Kipf, Ethan Fetaya, Kuan-Chieh Wang, Max Welling, and Richard Zemel.
\newblock Neural relational inference for interacting systems.
\newblock In \emph{International Conference on Machine Learning}, pages
  2688--2697. PMLR, 2018.

\bibitem[Yue et~al.(2014)Yue, Lucey, Carr, Bialkowski, and
  Matthews]{yue2014learning}
Yisong Yue, Patrick Lucey, Peter Carr, Alina Bialkowski, and Iain Matthews.
\newblock Learning fine-grained spatial models for dynamic sports play
  prediction.
\newblock In \emph{2014 IEEE international conference on data mining}, pages
  670--679. IEEE, 2014.

\bibitem[Sohn et~al.(2015)Sohn, Lee, and Yan]{sohn2015learning}
Kihyuk Sohn, Honglak Lee, and Xinchen Yan.
\newblock Learning structured output representation using deep conditional
  generative models.
\newblock \emph{Advances in neural information processing systems}, 28, 2015.

\bibitem[Kamra et~al.(2020)Kamra, Zhu, Trivedi, Zhang, and Liu]{kamra2020multi}
Nitin Kamra, Hao Zhu, Dweep~Kumarbhai Trivedi, Ming Zhang, and Yan Liu.
\newblock Multi-agent trajectory prediction with fuzzy query attention.
\newblock \emph{Advances in Neural Information Processing Systems},
  33:\penalty0 22530--22541, 2020.

\bibitem[Kingma and Welling(2013)]{kingma2013auto}
Diederik~P Kingma and Max Welling.
\newblock Auto-encoding variational bayes.
\newblock \emph{arXiv preprint arXiv:1312.6114}, 2013.

\bibitem[Gupta et~al.(2018)Gupta, Johnson, Fei-Fei, Savarese, and
  Alahi]{gupta2018social}
Agrim Gupta, Justin Johnson, Li~Fei-Fei, Silvio Savarese, and Alexandre Alahi.
\newblock Social gan: Socially acceptable trajectories with generative
  adversarial networks.
\newblock In \emph{Proceedings of the IEEE conference on computer vision and
  pattern recognition}, pages 2255--2264, 2018.

\bibitem[Paszke et~al.(2019)Paszke, Gross, Massa, Lerer, Bradbury, Chanan,
  Killeen, Lin, Gimelshein, Antiga, et~al.]{paszke2019pytorch}
Adam Paszke, Sam Gross, Francisco Massa, Adam Lerer, James Bradbury, Gregory
  Chanan, Trevor Killeen, Zeming Lin, Natalia Gimelshein, Luca Antiga, et~al.
\newblock Pytorch: An imperative style, high-performance deep learning library.
\newblock \emph{Advances in neural information processing systems}, 32, 2019.

\bibitem[Kingma and Ba(2014)]{kingma2014adam}
Diederik~P Kingma and Jimmy Ba.
\newblock Adam: A method for stochastic optimization.
\newblock \emph{arXiv preprint arXiv:1412.6980}, 2014.

\bibitem[Mohamed et~al.(2020)Mohamed, Qian, Elhoseiny, and
  Claudel]{mohamed2020social}
Abduallah Mohamed, Kun Qian, Mohamed Elhoseiny, and Christian Claudel.
\newblock Social-stgcnn: A social spatio-temporal graph convolutional neural
  network for human trajectory prediction.
\newblock In \emph{Proceedings of the IEEE/CVF Conference on Computer Vision
  and Pattern Recognition}, pages 14424--14432, 2020.

\bibitem[Australian Bureau~of Statistics(2021)]{labour}
2020 Australian Bureau~of Statistics.
\newblock Australian bureau of statistics. labour force, australia, dec 2020.,
  2021.
\newblock URL
  \url{https://www.abs.gov.au/statistics/labour/employment-and-unemployment/labour-force-australia/dec-2020}.
  Accessed on 01.18.2023.

\bibitem[Cuturi(2011)]{cuturi2011fast}
Marco Cuturi.
\newblock Fast global alignment kernels.
\newblock In \emph{Proceedings of the 28th international conference on machine
  learning (ICML-11)}, pages 929--936, 2011.

\bibitem[Dua and Graff(2017)]{duagraff}
D.~Dua and Graff.
\newblock Uci machine learning repository, 2017.
\newblock URL \url{http://archive.ics.uci.edu/ml}. Accessed on 01.18.2023.

\bibitem[Ben~Taieb and Koo(2019)]{ben2019regularized}
Souhaib Ben~Taieb and Bonsoo Koo.
\newblock Regularized regression for hierarchical forecasting without
  unbiasedness conditions.
\newblock In \emph{Proceedings of the 25th ACM SIGKDD International Conference
  on Knowledge Discovery \& Data Mining}, pages 1337--1347, 2019.

\bibitem[Salinas et~al.(2019)Salinas, Bohlke-Schneider, Callot, Medico, and
  Gasthaus]{salinas2019high}
David Salinas, Michael Bohlke-Schneider, Laurent Callot, Roberto Medico, and
  Jan Gasthaus.
\newblock High-dimensional multivariate forecasting with low-rank gaussian
  copula processes.
\newblock \emph{Advances in neural information processing systems}, 32, 2019.

\end{thebibliography}
